\documentclass[conference]{IEEEtran}
\IEEEoverridecommandlockouts
\usepackage{cite}
\usepackage{amsmath,amssymb,amsfonts}
\usepackage{graphicx}
\usepackage{textcomp}
\usepackage{xcolor}
\usepackage{ulem} 

\usepackage{amsthm}
\theoremstyle{remark}
\newtheorem{remark}{Remark}

\usepackage{bbding}
\usepackage{subcaption}

\usepackage{hyperref}
\hypersetup{
  colorlinks=true,
  linkcolor=blue,  
  citecolor=green,  
  urlcolor=blue    
}

\newtheorem{mytheorem}{Theorem}

\usepackage{algpseudocode}
\usepackage{algorithm}

\def\BibTeX{{\rm B\kern-.05em{\sc i\kern-.025em b}\kern-.08em
    T\kern-.1667em\lower.7ex\hbox{E}\kern-.125emX}}

\def\PredUnderNoise{\textsc{PredUnderNoise}}
\def\SampleUnderNoise{\textsc{SampleUnderNoise}}
\def\SampleFrom{\textsc{SampleUnderNoise}}
\def\LowerConfBound{\textsc{LowerConfBound}}
\def\UpperConfBound{\textsc{UpperConfBound}}
\def\CountPrediction{\textsc{CountPrediction}}

\begin{document}

\title{Fast Certification of Vision-Language Models Using Incremental Randomized Smoothing}

\author{\IEEEauthorblockN{Ashutosh Nirala}
\IEEEauthorblockA{
\textit{Iowa State University}\\
aknirala@iastate.edu}
\and
\IEEEauthorblockN{Ameya Joshi}
\IEEEauthorblockA{\textit{New York University} \\
ameya.joshi@nyu.edu}
\and
\IEEEauthorblockN{Soumik Sarkar}
\IEEEauthorblockA{\textit{Iowa State University} \\
soumiks@iastate.edu}
\and
\IEEEauthorblockN{Chinmay Hegde}
\IEEEauthorblockA{\textit{New York University} \\
chinmay.h@nyu.edu}
}

\maketitle

\begin{abstract}
A key benefit of deep vision-language models such as CLIP is that they enable zero-shot open vocabulary classification; the user has the ability to define novel class labels via natural language prompts at inference time. However, while CLIP-based zero-shot classifiers have demonstrated competitive performance across a range of domain shifts, they remain highly vulnerable to adversarial attacks. Therefore, ensuring the robustness of such models is crucial for their reliable deployment in the wild. 

In this work, we introduce Open Vocabulary Certification (OVC), a fast certification method designed for open-vocabulary models like CLIP via randomized smoothing techniques. Given a base ``training'' set of prompts and their corresponding certified CLIP classifiers, OVC relies on the observation that a classifier with a novel prompt can be viewed as a perturbed version of nearby classifiers in the base training set. Therefore, OVC can rapidly certify the novel classifier using a variation of incremental randomized smoothing. By using a caching trick, we achieve approximately two orders of magnitude acceleration in the certification process for novel prompts. To achieve further (heuristic) speedups, OVC approximates the embedding space at a given input using a multivariate normal  distribution bypassing the need for sampling via forward passes through the vision backbone. We demonstrate the effectiveness of OVC on through experimental evaluation using multiple vision-language backbones on the CIFAR-10 and ImageNet test datasets.
\end{abstract}



\section{Introduction}

\subsection{Motivation}
Deep learning systems have achieved state-of-the-art performance in various domains, including computer vision \cite{karpathy2014learned}, speech \cite{radford2023robust}, and more \cite{lakhani2017deep,devlin2018bert,sauer2022stylegan}, occasionally surpassing human capabilities \cite{hannun2021history}. Recently, significant progress has been made towards building \emph{vision-language} models that are trained via self-supervision on traw unlabeled datasets of paired images and their text captions scraped from the internet. This has led to the development of \emph{open vocabulary} models such as CLIP\cite{radford2021learning}, OpenCLIP \cite{cherti2023reproducible} and OSCAR\cite{li2020oscar}. These models excel at \emph{zero-shot} image classification: a user has the ability to specify novel class labels using natural language prompts at inference time.

However, adversarial attacks have consistently posed a significant challenge for computer vision  \cite{szegedy2014intriguing,goodfellow2018defense,joshi2019semantic} and other deep learning \cite{lee2020spatiotemporally} systems. A meticulously designed perturbation, imperceptible to humans, can severely impair their performance. 
This issue has raised substantial concerns regarding the deployment of such systems in safety-critical applications. We show below that zero-shot vision language models are especially vulnerable to such attacks (even more so than standard models based on supervised training). In response to the emergence of such adversarial attacks, various defense methods have been proposed.  
The majority of these methods are based on Adversarial Training~\cite{madry2018towards,wang2019improving,zhang2019theoretically,wu2020adversarial} but they lack robustness guarantees, leaving room for potential accuracy reduction through novel attacks. 
\subsection{Need for Model Certification}
The absence of robustness guarantees in safety-critical systems is concerning and limits their broader applicability. \cite{athalye2018obfuscated} highlighted that many defenses provide a false sense of security by obfuscating gradients. Additionally, since adversarial attacks can transfer across networks \cite{szegedy2014intriguing}, crafting an adversary on a surrogate model can compromise the deployed model. Thus, merely relying on empirical robustness evaluations may not suffice for reliable deployment.
Consequently, a parallel line of work towards development of certified robustness has emerged. These ensure that the model's output for a given input provably (or certifiably) remains unchanged within a certain neighborhood, $R$, of the input.  
For instance, if a model is certified for an input up to a radius $R$ in $\ell_p$, it guarantees that any adversarial attack, including FGSM \cite{szegedy2014intriguing}, PGD \cite{madry2018towards}, AA \cite{croce2020reliable}, Square \cite{andriushchenko2020square}, RayS \cite{chen2020rays}, or others, will not alter the model's prediction if the perturbation is $< R$ in $\ell_2$.

Among the various certification methods proposed, such as  \cite{katz2017reluplex,wong2018provable,salman2019provably,cohen2019certified}, those based on randomized smoothing stand out for their scalability, i.e., they can be feasibly applied to larger networks. At a high level, these methods rely on taking a base (deep) classifier (say $f$) and ``smoothing'' it by convolving with a probability density function (say $h$), such as a Gaussian function. This process yields bounds on the Lipschitz constant of the smoothed model $f \star h$, giving certificates of correctness within a certain perturbation radius around a given input. However, certification speeds are still rather slow: in practice, such a convolution is achieved by adding sampled noise to the input, performing a forward pass to obtain a class prediction, and averaging the predictions over (hundreds of) thousands of samples. This poses a challenge particularly in the context of vision-language models; since prompts can be typically constructed by the user at inference time, \emph{quickly} certifying the constructed classifier becomes paramount.

\subsection{Our Contributions}

In this paper, we introduce and validate a  framework for certifying zero-shot vision-language classifiers using randomized smoothing. We call our method \emph{Open Vocabulary Certification} (or OVC).   

Our OVC framework is based on the following intuition. Suppose we start with a large set of image classifiers based on known (``training'') prompts, and pre-compute their corresponding certificates for a given set of input images. Now, for a given input image, one would expect a pair of similar prompts (as measured with respect to the text embedding space) to lead to the same class prediction; moreover, one would expect \emph{perturbations} of an input image to (mostly) lead to the same class prediction. Therefore, if a novel (``test'') prompt is nearby one of the prompts in the known set, then we can simply retrieve the certificate produced at the pre-computation stage.  Errors might occur if the prompts are too far away, or if the confidence (logit) levels are too close; for such cases, we can certify the model for that input from scratch. We note that this idea is reminiscent of Incremental Randomized Smoothing (IRS), recently proposed in \cite{ugare2023incremental}. 

However, applying IRS directly to CLIP-style classification presents unique challenges. First, to reliably work, IRS traditionally assumes minimal output deviation, (not exceeding $1\%$) across different models—equivalent to prompts in our context. Therefore we adapt IRS; our adapted IRS version brings significant improvements by speeding up the certification time for a novel prompt, capitalizing on insights derived from existing prompts. For example, for ImageNet for $\sigma = 0.25$, our modified IRS boosts the certification time by $1.32\times$ for CLIP with a Resnet-50 backbone.

Second, we leverage the following property of CLIP: even though prompts are modified, the embeddings for a given input image remain unchanged. Given that randomized smoothing necessitates repeated passes of the input with added Gaussian noise (hundreds of) thousands of times, we can achieve substantial acceleration by implementing an embedding caching strategy. By caching the input image embeddings during the certification of existing or previous prompts, we achieved \emph{two orders of magnitude} acceleration in the certification process, albeit with increased storage requirements.

Third, in order to alleviate storage costs due to caching the embeddings, we instead perform a fast (but heuristic) approximation method by fitting a multivariate Gaussian (MVN) distribution. Leveraging the multivariate normal approximation offers notable advantages: it significantly reduces the computational cost of sampling compared to using CLIP directly, and it eliminates the need to load embeddings from disk, further expediting the certification process for novel prompts.

Note that this last step is heuristic and does not lead to provable certificates. Instead, we provide an extensive empirical analysis comparing perturbation radii obtained with and without the MVN approximation. Particularly for larger radii, there are instances where the obtained radius slightly exceeds the certified radius obtained through randomized smoothing. Empirically, we find that by merely reducing the probability of the top-most prediction by a mere $1\%$, our method is reliable: the calculated radius almost always undershoots the actual radius, as obtained without MVN approximation, for both CIFAR-10 and ImageNet datasets. Quantifying the error in approximating the pre-logit space using an MVN remains a valuable avenue for future research.

\subsection{Summary and Organization}
To summarize, our contributions in this paper are as follows:
\begin{enumerate}
    \item \textbf{Open Vocabulary Certification (OVC)}: We introduce the concept of Open Vocabulary Certification (OVC). This approach harnesses certificates pre-computed for an existing set of prompts in order to expedite the certification of new prompts efficiently. 
    \item \textbf{Methods for OVC}: We present both exact and heuristic methods for fast Open Vocabulary Certification, including adaptations of the existing IRS method, as Modified-IRS, to suit the OVC framework. Specifically the three methods are: Modified-IRS, Cached-OVC and MVN-OVC.
    \item \textbf{Empirical Validation:} We validate our approachy through extensive certification experiments conducted on CLIP (RN50 and ViT-B-32) and OpenCLIP (ViT-B-32) across two standard image classification datasets, namely ImageNet and CIFAR-10.
\end{enumerate}

The remainder of this paper is structured as follows. In the next section, we delve into the background and related work. Subsequently, we detail our methods for Open Vocabulary Certification. This is followed by the experiments section, where we showcase the effectiveness of our approach on ImageNet and CIFAR-10 datasets. Finally, we conclude the paper with discussions on our findings and prospects for future work.

\section{Background and Related Work}
In this section, we first give a brief preliminary about certification problem and the notations used. Followed by that we talk briefly about adversarial attacks and defenses. Then we discuss the need of certification. In related work we discuss the randomized smoothing as introduced by Cohen et al., \cite{cohen2019certified}, along with few other varaitions.

\subsection{Preliminaries}
We first introduce some basic notation. The goal of an ideal classifier $f(\cdot)$ is to correctly assign an input point $x$ to its correct class $y$. Specifically, we represent multi-class classifiers which assign the given input to one of $K$ classes by outputting logits (i.e., real numbers) $f_i(x)$ and setting the predicted label $y_p$ for the classifier as:
\begin{equation}
    y_p = \arg\max_{i \in [K]} f_i(x)
\end{equation}
Since classifiers are susceptible to adversarial attacks, we are interested in calculating a radius of certification, $R$, such that for all points within a ball of radius $R$ around the input $x$, the classifier does not change its output. The ball is typically defined in terms of its $p$-norm i.e., 
$$\|x - x'\|_p \leq R.$$ 
Throughout this paper we work with $\ell_2$ certificates, i.e., $p = 2$. 

Computing a tight estimate for the radius of certification, $R$, can be intractable for classifiers $f$ that are implemented by practical deep neural networks. For such networks, an alternative is to use randomized smoothing (RS), which we describe below.

\subsection{Adversarial attacks and defenses}
Adversarial attacks are broadly classified into white and black box attacks. In white box attacks, the adversary has complete access to the model including its gradient. While in black-box adversary has only limited access. Further an attack can be targeted or untargeted. In the targeted attack, adversary tried to perturb the input such that it gets misclassified to a given target class, while in the untargeted case, the goal is to cause misclassification irrespective of the specific choice of target.

FGSM \cite{goodfellow2014explaining} and its iterative variant, PGD \cite{kurakin2018adversarial,madry2018towards} are the most widely used white-box attacks. AutoPGD \cite{croce2020reliable}, FAB \cite{croce2020minimally}, and C\&W \cite{carlini2017towards} attack are some popular variations of PGD. In most real world scenarios, an adversary rarely has internal access to the deployed model. 
In such cases, they can either resort to transfer attack i.e., use a white-box attack on another accessible model, and transfer this to the deployed model. If the attacker has access to the score of the model for various classes,  they can use this score as a guide to find adversarial perturbation directions using random walks. The SQUARE attack \cite{andriushchenko2020square} is a popular choice for such scenarios. If the attacker only has query access to the predicted labels, they can resort to hard-label black-box attacks like RayS \cite{chen2020rays}, SPSA \cite{uesato2018adversarial}, and HopSkipJump \cite{chen2020hopskipjumpattack}.

\textbf{Empirical Defenses:} Defenses based on adversarial training, where the adversarial samples are generated and incorporated during the training of the model, SAT \cite{madry2018towards} has been most successful. TRADES \cite{zhang2019theoretically} is a notable variation of SAT. Few other AT methods are MART \cite{wang2019improving}, HE \cite{pang2020boosting,fakorede2023improving}, AWP \cite{wu2020adversarial,yu2022robust} and \cite{esfandiari2021fast,havens2018online} among others. 

\textbf{Certified Defenses:} While a handful of the above empirical defenses work across different datasets, none come with any guarantees about their robustness. In fact several of the previous empirical defenses were later broken by stronger attacks. 
Consequently, a parallel line of work towards the development of certified robustness has emerged. We primarily classify these into methods employing Randomized Smoothing (RS) and those that don't.

The ones in the latter category establish an upper bound on the certification radius by establishing the bound at the input layer and limit it by propagating it across each layer, using linear, quadratic, convex or integer-mixed programming. They include methods like  Carlini et al. \cite{carlini2017ground}, Huang et al. \cite{huang2017safety}, Katz et al. \cite{katz2017reluplex,katz2017towards} which includes Reluplex, Weng et al. \cite{weng2018towards}. Wong and Kolter \cite{wong2018provable} and Raghunathan et al. \cite{raghunathan2018certified,raghunathan2018semidefinite}. These methods are computationally very expensive, and therefore unfeasible for larger networks.

Early nethods based on RS include Cohen et al. \cite{cohen2019certified}, Lecuyer et al. \cite{lecuyer2019certified} and Salman et al. \cite{salman2019provably}; all these provide $\ell_2$ robustness certificates. At a high level, the network is convolved with a Gaussian noise distribution to smooth its functionality. \cite{tengl1,levine2021improved} presents methods using Laplacian smoothing in oder to provides certificates for $\ell_1$ and Wasserstein metrics. While approaches to select the distribution for various classes of adversarial attacks has been presented (such as Yang et al \cite{yang2020randomized}) certificates for perturbations other than $\ell_2$-norm balls have an $\Omega(d^{-1/2})$ dependence, and therefore are too small to be useful. 

Typically, RS defenses provide certificate radii which are smaller than those provided by empirical defenses. Various methods like MACER \cite{zhai2020macer}, Alfarra et al, \cite{alfarra2022data}, and Jeong et al. \cite{jeong2021smoothmix} has been proposed to bridge this gap. However they all involve re-training large-scale models with different objectives, and are out of scope for this work. Next, we formally describe tools which are most closely related to our work.


\subsection{Randomized smoothing basics}

\begin{table*}[!tbp]
\caption{Categorization of certification methods}
\begin{center}
\begin{tabular}{|l|c|c|c|c|}
\hline
\textbf{Methods} & \textbf{Type}
                 & Scalable
                 & Reusable
                 & Speedup$^{\mathrm{b}}$\\
\hline
Non-smoothing$^{\mathrm{a}}$ & Exact  & \textcolor{red}{\XSolid} & \textcolor{red}{\XSolid} & N.A.\\
Cohen et al, \cite{cohen2019certified}, Salman et al, \cite{salman2019provably} & Probabilistic  & \textcolor{green}{\Checkmark} & \textcolor{red}{\XSolid} & 1x (baseline)\\

Modified-IRS (Ours) & Probabilistic  & \textcolor{green}{\Checkmark} & \textcolor{green}{\Checkmark} & 0.94x - 1.68x\\
OVC (Ours) & Probabilistic & \textcolor{green}{\Checkmark} & \textcolor{green}{\Checkmark} & 46x\\
MVN-OVC (Ours) & Approximate  & \textcolor{green}{\Checkmark} & \textcolor{green}{\Checkmark} & 137x\\
\hline
\multicolumn{5}{l}{$^{\mathrm{a}}$These include linear/semidefinite programming methods such as  \cite{katz2017reluplex,carlini2017ground,huang2017safety,katz2017towards,weng2018towards,wong2018provable,raghunathan2018certified,raghunathan2018semidefinite}.}\\
\multicolumn{5}{l}{$^{\mathrm{b}}$The speedup is shown for ImageNet for different values of noise $\sigma$ using CLIP-RN50 as backbone.}\\
\multicolumn{5}{l}{For our methods, the speedups are achieved for novel prompts by reusing information from existing prompt certifications.}

\end{tabular}
\label{tab1}
\end{center}
\end{table*}

Our OVC framework can be viewed as an extension of randomized smoothing as described in Cohen et al, \cite{cohen2019certified}. Throughout the paper we refer to RS as the ``standard'' approach for certification. Here, we restate their algorithm and main theorem; we later adapt both these elements when introducing our methodology.

\textbf{Randomized Smoothing:} The high level idea in RS is to consider a surrogate network $g$ which is a convolved (smoothed) version of the original/base network $f$ with a Gaussian distribution. Mathematically, the prediction for $g$ is the most likely class returned by $f$ for input $x$, when the input is perturbed by isotropic Gaussian noise:
\begin{equation}
\begin{aligned}
    &g(x) = \arg \max_{c\in [K]} P( f(x + \epsilon) = c) \\
    &\text{where } \epsilon \sim \mathcal{N}(0, \sigma^2I) \, .
\end{aligned}
\label{eq:smooth_clsf}
\end{equation}
Curiously, this operation leads to provable certificates. For the smoothed classifier $g$ defined in eq. \ref{eq:smooth_clsf}, we have the following theoretical guarantee.
\begin{mytheorem}
   Let $x$ be an input. Let $\underline{p_A}$, $\overline{p_B}$ $\in[0, 1]$, where they represent the lower and upper bounds on its most probable class and runner-up class respectively, satisfy:
   \begin{equation}
       P(f(x + \epsilon) = c_A) \geq \underline{p_A} \geq \overline{p_B} \geq \max_{c \neq c_A} P(f(x + \epsilon) = c)
   \end{equation}
   
   Then, necessarily $g(x + \delta) = c_A$ for all $\|\delta\|_2 < R$, where
   \begin{equation}
       R = \frac{\sigma}{2}(\Phi^{-1}(\underline{p_A}) - \Phi^{-1}(\overline{p_B})) 
   \end{equation}

   Here, $\Phi^{-1}$ denotes the inverse of the standard Gaussian CDF.
   \label{thm:cohen}
\end{mytheorem}

Computing the exact probability for each class requires an expectation over the (complicated) distribution induced by a pushforward of a Gaussian through a general neural network and is not tractable; therefore, most RS methods resort to Monte Carlo sampling. Practically, we set $\overline{p_B} = 1 - \underline{p_A}$, and declare $R = \sigma \cdot \Phi^{-1}(p_A)$. We abstain from certifying and making prediction if $p_A < \frac{1}{2}$. Overall, this approach is termed the CERTIFY algorithm as presented by Cohen et al \cite{cohen2019certified}, and is described in pseudocode form in Alg \ref{alg:Cohen_algo}. It uses the following functions:
\begin{itemize}
    \item \SampleUnderNoise$(f, x, n, \alpha)$: Returns the count for each predicted class for the input $x$ for the base network $f$ when the input is perturbed by Gaussian noise with standard deviation $\sigma$. 
    \item \LowerConfBound$(p, n, 1 - \alpha)$: Returns a lower bound on probability $p$ when sampled with $n$ samples, with confidence $1 - \alpha$ via the Clopper-Pearson Lemma\cite{clopper1934use}.
\end{itemize}

\begin{algorithm}[htbp]
\caption{Randomized smoothing certification algorithm, $CERTIFY(f, \sigma, x, n_0, n, \alpha)$ as presented by Cohen et al., \cite{cohen2019certified}. We call this algorithm from our Modified-IRS algorithm}\label{alg:Cohen_algo}
\textbf{Inputs:}
\begin{itemize}
    \item $f$: Given base neural network.
    \item $\sigma$: Std-dev of Gaussian noise used for certification.
    \item $x$: Input.
    \item $n_0$: \# samples to predict the top class.
    \item $n$: \# samples for computing $\underline{p_A}$.
    \item $\alpha$: Confidence parameter.
\end{itemize}

\textbf{Output}
\begin{itemize}
    \item Predicted class $c_A$ for input $x$, along with certified radius $R$ or ABSTAIN.
\end{itemize}

\hrule
\begin{algorithmic}[1]
\State $counts0$ $\leftarrow$ \SampleUnderNoise$(f, x, n_0, \alpha)$.
\State $\hat{c}_A$ $\leftarrow$ top index in $counts0$.
\State $counts$  $\leftarrow$ \SampleUnderNoise$(f, x, n, \alpha)$.
\State $\underline{p_A}$  $\leftarrow$ \LowerConfBound$(counts[\hat{c}_A], n, 1 - \alpha)$.
\If{$\underline{p_A} > \frac{1}{2}$}
    \State \textbf{return} Class: $\hat{c}_A$, Radius: $\sigma \cdot \Phi^{-1}(p_A)$.
\Else
    \State \textbf{return} ABSTAIN.
\EndIf
\end{algorithmic}
\end{algorithm}
First it determines the majority class using $n_0$ samples. Then it estimates $\underline{p_A}$ using $n$ samples. Finally based on $p_A$, it either returns the prediced class and certification radius or abstains from doing so.

\textbf{Incremental Randomized Smoothing: }In very recent work \cite{ugare2023incremental}, the authors propose an adaptation of randomized smoothing called incremental randomized smoothing (IRS) to produce certificates for a model which is obtained by quantizing (or pruning) a pre-certified model. They observe that, in their case, the predictions by the original and the derived model do not differ much. Specifically, they found that, under Gaussian noise, the prediction error never exceeded more than $1\%$. They further point out that such small errors can be estimated using existing binomial proportion estimation techniques using fewer perturbed samples. Therefore, by leveraging knowledge of the pre-computed certificates, IRS leads to faster certification of the derived models.

In essence, given a model's prediction under Gaussian noise, IRS determines the prediction for the modified model under the same noise conditions. This is achieved by caching the seeds used for generating the Gaussian noise. Then, using binomial confidence upper limit using Clopper and Pearson \cite{clopper1934use} method, they probabilistically assess the prediction difference. With a typically small probability, a reliable estimate is obtained using fewer Gaussian perturbations for the modified network, such as 10K instead of 100K samples. If the difference in $p_A$ is $\zeta$, and $p_A - \zeta > 0.5$, the certification radius is confirmed to be at least $> \sigma \Phi^{-1}(p_A - \zeta)$, according to \cite{cohen2019certified}. The IRS algorithm is detailed in Appendix \ref{app:IRS} (Algorithm \ref{alg:IRS} outlines the main IRS algorithm, and \ref{alg:ZetaEst} is the subroutine for estimating error differences). We will borrow this intuition while developing our OVC framework for Modified-IRS. 

\subsection{Zero-shot Vision-Language  Classifiers}
In 2021, OpenAI released CLIP \cite{radford2021learning}, introducing a new paradigm in image classification called: \textit{Zero-shot Vision-Language  Classifiers}. Since its release, CLIP has garnered over $10,000$ citations, indicating its widespread adoption as a backbone in image classification systems. Recently OpenCLIP \cite{cherti2023reproducible} have investigated scaling laws for CLIP by training on public LAION dataset. These classifiers are trained on vast collections of internet-sourced image and caption pairs. During training, images and the text from captions are encoded using separate vision and text encoders. The goal is to align the two encodings (embeddings) for each pair, which means enhancing the dot product value between the embeddings of a pair. The training loss penalizes misalignment with disparate caption embeddings and rewards alignment with corresponding image-caption pairs, using large batch sizes.

Post-training, the image and text encoders produce aligned embeddings for corresponding images and captions. For classification, rather than using captions directly, one designs prompts for each image class. These prompts describe the image class, such as \textit{"a picture of a ship"} for the class \textit{"ship"}. The target class is determined by the highest alignment, or dot product value, between the input image embeddings and the prompt embeddings. We have provided more details for CLIP in Appendix \ref{sec:clipOverview}.

\begin{figure*}[!htbp]
    \centering
    \includegraphics[width=0.82\textwidth]{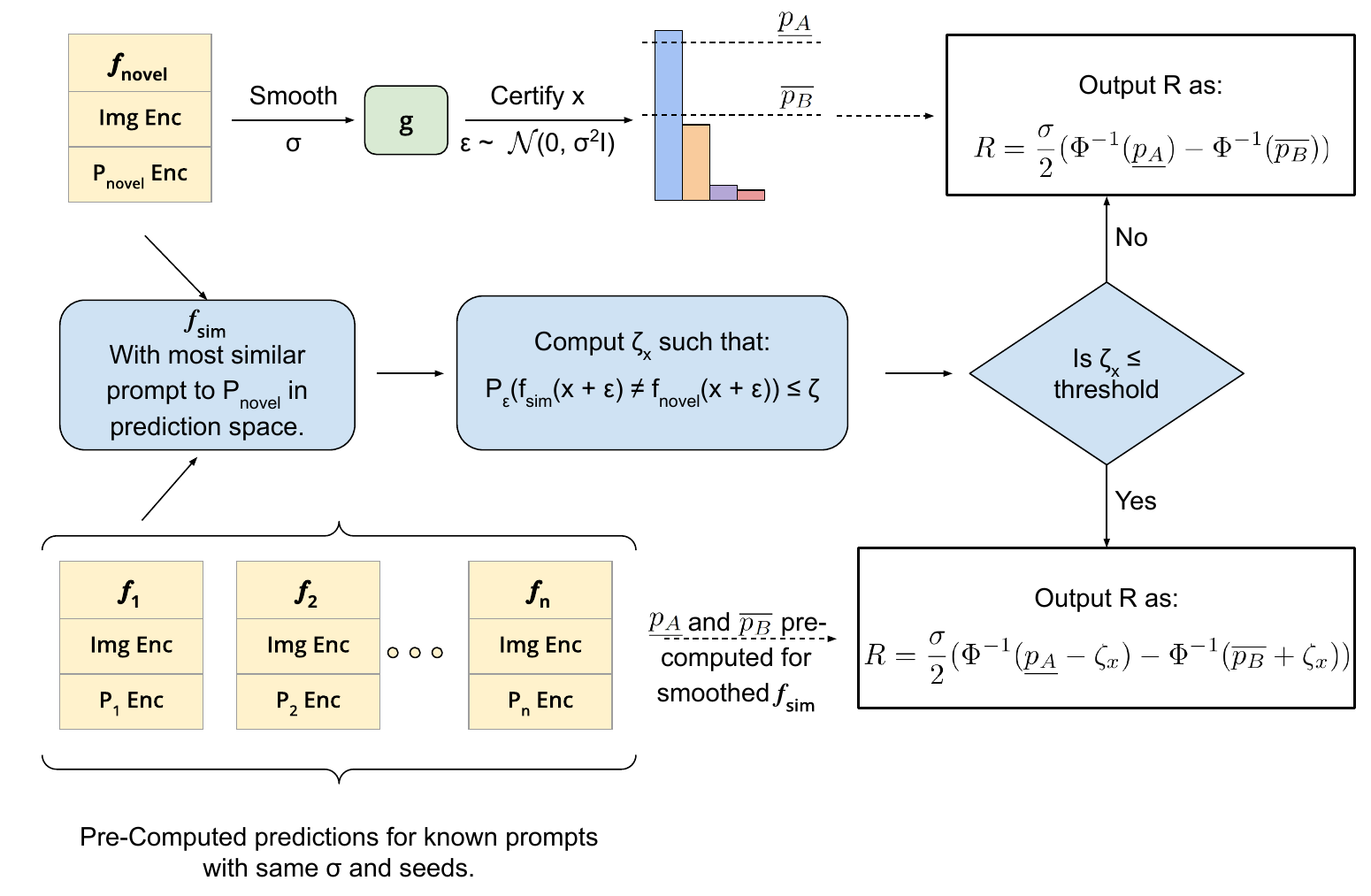}
    \caption{Workflow of OVC. For a given prompt, using relatively few samples, we find a prompt (out of prompts whose prediction is known)  which is most similar in prediction to the given prompt. If the difference in prediction is below certain threshold, we certify using the information from existing prompt, saving time.}
\label{fig:ovc_flow}
\end{figure*}

\begin{figure*}[!htbp]
\begin{subfigure}{0.3\textwidth}  
        \centering
        \includegraphics[width=\textwidth]{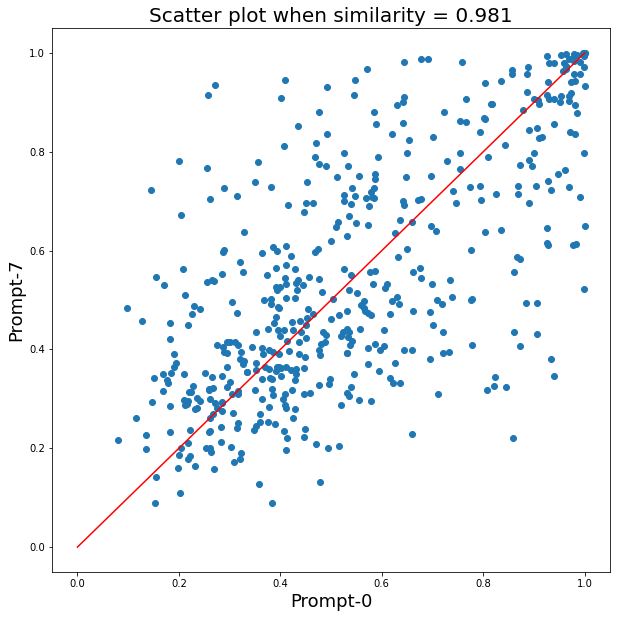}
        \caption{Similar prompts}
    \end{subfigure}%
    \hfill
    \begin{subfigure}{0.3\textwidth}  
        \centering
        \includegraphics[width=\textwidth]{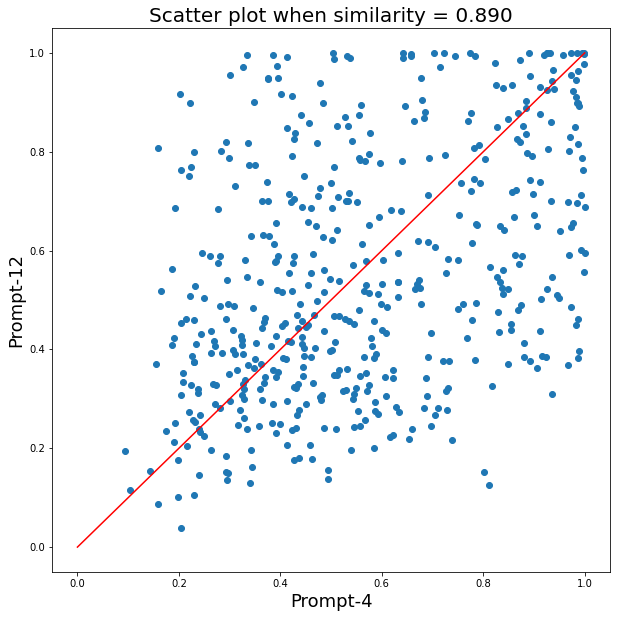}
        \caption{Two random prompts}
    \end{subfigure}%
    \hfill
    \begin{subfigure}{0.3\textwidth}  
        \centering
        \includegraphics[width=\textwidth]{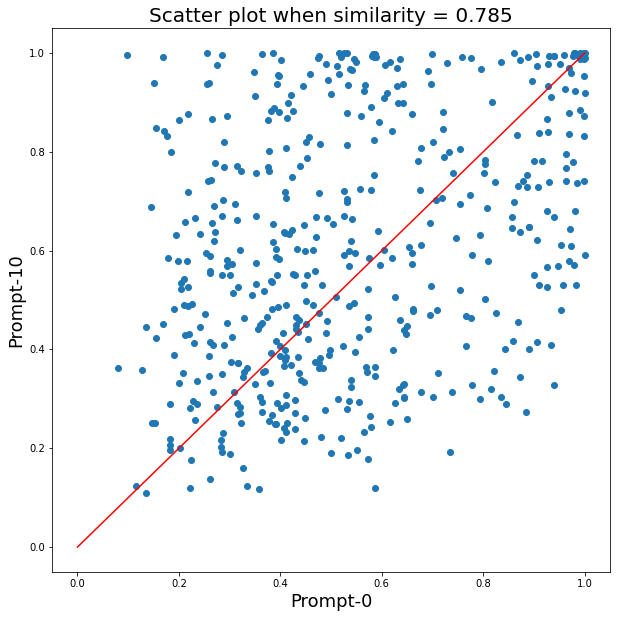}
        \caption{Dissimilar prompt}
    \end{subfigure}

    \caption{Scatter plot showing the relationship between probability of top-most class, $p_A$, for prompts with varying degree of similarity. Even when the two prompts are very close in cosine similarity, Fig. (a), they vary widely on the probability for the top-most class, indicating that IRS \cite{ugare2023incremental} can not be applied directly for OVC. 
    All the certificates have been computed for CLIP-RN50 on ImageNet with $\sigma = 0.25$}
\label{fig:not_sim}
\end{figure*}

\section{Method: Open Vocabulary Certification}
Our goal is to devise a fast certification method for zero-shot vision-language models. The uniqueness of this setting is that the full classifier is not known during training; in CLIP, for example, the classifier varies according to the choice of prompt at inference time. The key challenge is to come up with an efficient certification method in this dynamic setting where we can quickly produce certificates for a novel prompt; we achieve this using information obtained while certifying existing prompts.

\begin{figure}[!htbp]
    \centering
    \includegraphics[width=0.45\textwidth]{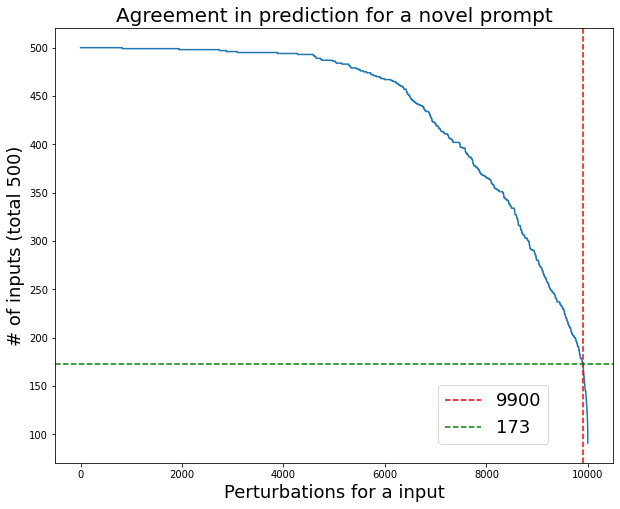}
    \caption{Plot shows agreement in predictions for the input, for a novel prompt with predictions made by existing prompts. For $10K$ random perturbations with $\sigma=0.25$, among $500$ input samples tested, we found that for about $30\%$, there is at least one prompt for which disagreement is $< 1\%$.
}
\label{fig:simPrompt}
\end{figure}

\subsection{Modified-IRS}
As a first attempt, we directly apply a  version of IRS \cite{ugare2023incremental} for our problem. 
The key assumption in IRS is that the two networks do not differ in their prediction by 
much.
In our case, we hypothesize that if the text embeddings for a pair of prompts are similar . than their certificates will also be similar. 

We tested this hypothesis for the ImageNet benchmark. We consider the set of 80 prompts suggested in the official CLIP repository (\url{https://github.com/openai/CLIP/blob/main/data/prompts.md}) for ImageNet. For our setting, we randomly divided the prompts into 70 known (``train'') prompts and 10 unknown (``test'') prompts. We assume that, for all train prompts, we have the certificate as well as ancillary  information (like seeds used for Gaussian noise) already calculated and available. Our goal is to use this information to certify the classifier for a novel test prompt in relatively less time than it would take to certify from scratch.

Let us apply IRS to this setting in a straightforward manner. Among the known classifiers, i.e., train prompts, we need to identify the one which is most similar to the novel prompt. 
To measure similarity, we concatenate the embeddings of the prompts (using the CLIP text encoder) for all 1000 ImageNet classes, and use this vector representation to compute cosine similarities. Somewhat surprisingly, we find that for the pair prompts which are most similar (with cosine similarity $> 0.98$) the top-most class probability $p_A$ also varies widely. This is  illustrated in Fig \ref{fig:not_sim}. Since the difference in the predicted $p_A$, for the majority of samples, is mostly greater than $1\%$, we can not apply IRS directly out of the box.

However, a simple modification of this idea is successful. Given an input, we may search for the train prompt which is most similar in its prediction \emph{for that specific input}. Like IRS, we can establish this using only few perturbations. We tested this for our novel prompts against $70$ known prompts. For $10K$ perturbations with $\sigma=0.25$, among $500$ input samples tested, we found that for more than $30\%$ there is at least one prompt for which the probability of disagreement is $<1\%$. We plot the agreement in Fig \ref{fig:simPrompt}. 
For any given input, if we find an existing prompt where the disagreement is minimal (less than $1\%$ as considered in the original IRS method), then IRS can be effectively applied for that input. 
If however there is no known prompt for which the difference in prediction is small enough, we resort to full certification using Algo \ref{alg:Cohen_algo}, i.e., using a larger number of perturbed samples. We call this method \emph{Modified-IRS} and summarize our algorithm in Algo \ref{alg:m_IRS}.

\begin{algorithm}[htbp]
\caption{Modified-IRS$(f, \sigma, x, n_0, n_p, n, \alpha, \alpha_\zeta, \mathcal{C}_f, \gamma)$}
\label{alg:m_IRS}
\textbf{Inputs:}
\begin{itemize}
    \item $f$: Given base vision-language model:
        \begin{itemize}
            \item $f_{im}$: Encodes image.
            \item $f_p$: Encodes prompt.
            \item $prompt$: Prompt for all classes.
        \end{itemize}
    \item $\sigma$: Std-dev of Gaussian noise used for certification.
    \item $x$: Input.
    \item $n_0$: \# samples to predict the top class.
    \item $n_p$: \# samples to find the most similar prompt in prediction.
    \item $n$: \# samples for computing $\underline{p_A}$.
    \item $\alpha, \alpha_\zeta$: Confidence parameters.
    \item $\mathcal{C}_f$: Cache storing information while certifying known prompts for input $x$. See text for details.
    \item $\gamma$: maximum allowed difference in prediction to use IRS.
\end{itemize}
\textbf{Output}
\begin{itemize}
    \item Predicted class $c_A$ for input $x$, along with certified radius $R$ with $1 -\alpha -\alpha_\zeta$ confidence or ABSTAIN.
\end{itemize}
\hrule
\begin{algorithmic}[1]
\State $pred\_p$ $\leftarrow$ \PredUnderNoise$(f, x, n_p, \alpha, \mathcal{C}_f[seeds])$.
\State $sim_p$ $\leftarrow$ Most similar prompt as per $pred\_p$ as stored in $ \mathcal{C}_f$.
\State $diff$ $\leftarrow$ count of ($\mathcal{C}_f[sim_p][pred][:n_p] \neq pred\_p$).
\If{$diff/n_p > \gamma$} 
    \State \textbf{return} $CERTIFY(f, \sigma, x, n_0, n, \alpha + \alpha_\zeta)$.
\Else               
    \State $\zeta_x$ $\leftarrow$  \UpperConfBound$(diff, n_p, 1 - \alpha_\zeta)$.
    \State $\underline{p_A}$ $\leftarrow$ $\mathcal{C}_f[sim_p][p_A]$
    \If ( $\underline{p_A} - \zeta_x > \frac{1}{2}$)
        \State \textbf{return} Class: $\mathcal{C}_f[sim_p][c_A]$, Radius: $\sigma \cdot \Phi^{-1}(\underline{p_A} - \zeta_x)$.
    \Else
        \State \textbf{return} ABSTAIN.
    \EndIf
\EndIf
\end{algorithmic}
\end{algorithm}

The algorithm leverages pre-computed information about the classiifers corresponding to train prompts stored in a cache $\mathcal{C}_f$ with the following fields. Note that this cache is specific to input $x$.
\begin{itemize}
    \item $\mathcal{C}_f[seeds]$: Seeds used for certifying the known prompts.
    \item $\mathcal{C}_f[prompt][pred][:n_p]$: Returns the first $n_p$ prediction for $prompt$.
    \item $\mathcal{C}_f[prompt][p_A], \mathcal{C}_f[prompt][c_A]$: Returns $\underline{p_A}$ and $\hat{c}_A$ for the given input $x$ and $prompt$.
\end{itemize}

The algorithm also makes use of the following new functions:
\begin{itemize}
    \item \PredUnderNoise$(f, x, n, \alpha, seeds)$: This is similar to the \SampleUnderNoise$(\dots)$ function, but instead of returning the prediction count for each class it simply returns the prediction for all the $n$ perturbations. It uses the $seeds$ passed to it to sample the Gaussian noise.
    \item \UpperConfBound$(diff, n_p, 1 - \alpha_\zeta)$: Like \LowerConfBound$(\dots)$, but it returns an upper bound.
\end{itemize}

As noted in our results below, we show a savings of approximately $30\%$ compute time when we use Modified-IRS for certifying novel prompts, compared to applying RS from scratch.

Similar to IRS, we obtain following theoretical result for Modified-IRS:
\begin{mytheorem}
   Let $f_{novel}$ be a zero-shot classifier defined using a novel prompt. Suppose there is an existing train prompt $sim$, with corresponding classifier $f_{sim}$ such that for a given $x$, $P_\epsilon(f_{sim}(x + \epsilon) \neq f_{novel}(x + \epsilon)) \leq \zeta_x$, and $f_{sim}$ satisfies 
   $$\begin{aligned}
   P_\epsilon(f_{sim}(x + \epsilon) = c_A) &\geq \underline{p_A} \geq \overline{p_B} \\ & \geq max_{c \neq c_A}P_\epsilon(f_{sim}(x + \epsilon) = c)
   \end{aligned}
   $$ and $\underline{p_A} - \zeta_x \geq \overline{p_B} + \zeta_x$. Then for the smoothed classifier obtained for the novel prompt,  $g_{novel}$, we necessarily have $g_{novel}(x + \delta) = c_A$ for all $\|\delta\|_2 < R$, where: 
   
   \begin{equation}
       R = \frac{\sigma}{2}(\Phi^{-1}(\underline{p_A} - \zeta_x) - \Phi^{-1}(\overline{p_B} + \zeta_x)) 
   \end{equation}

   Here, $\Phi^{-1}$ denotes the inverse of the standard Gaussian CDF.
   \label{thm:m_irs}
\end{mytheorem}

The proof is same as the proof for IRS algorithm \cite{ugare2023incremental}, where we replace $f$ with $f_{sim}$ and specialize to a specific input. We omit this proof for brevity.

\textbf{Highlighting the Difference Between IRS and Modified-IRS: } Below, we outline the key differences between IRS \cite{ugare2023incremental} and Modified-IRS (our method):
\begin{itemize}
    \item In the IRS setting, there is only one base model. In contrast, Modified-IRS deals with multiple prompts, thus involving several base models. We refer to these prompts as the known training set.
    \begin{itemize} 
        \item Consequently, Modified-IRS necessitates identifying a prompt from the test set that closely resembles the one being certified, where similarity is defined by the consistency of predictions for a given input under Gaussian noise.
    \end{itemize}
    \item In the IRS setting, the base model and the model to be certified never differ in their predictions by more than $1\%$ probability. However, in Modified-IRS, only about $30\%$ of samples (when Gaussian noise, $\sigma = 0.25$) match this criterion. For the remaining samples, complete certification using $100K$ perturbations is necessary. Additionally, as the noise level (i.e., $\sigma$) increases, the proportion of agreeing samples decreases, limiting the speedup provided by Modified-IRS at higher noise levels.
\end{itemize}

\subsection{Caching embeddings (OVC)}
For open vocabulary models like CLIP, there are two steps involved in image classification. First, the (image) embeddings for both input, and the (text) embeddings for the prompt for each class is calculated. Then, the logit for each class is calculated via a dot product between the image embedding and corresponding prompt embedding. We notice that for a novel prompt, the embedding for the input image does not change. Therefore, we can further improve IRS performance by caching all image embeddings. We call this improved version \emph{OVC} (which is our main algorithm) and describe it in pseudocode form in Alg. \ref{alg:ovc}.

\begin{algorithm}[htbp]
\caption{OVC$ (f, \sigma, x, n_0, n, \alpha,\mathcal{C}_f)$}
\label{alg:ovc}
\textbf{Inputs:}
\begin{itemize}
    \item $f$: Given base vision-language model:
        \begin{itemize}
            \item $f_{im}$: Encodes image.
            \item $f_p$: Encodes prompt.
            \item $prompt$: Prompt for all classes.
        \end{itemize}
    \item $\sigma$: Std-dev of Gaussian noise used for certification.
    \item $x$: Input.
    \item $n_0$: \# Samples to predict the top class.
    \item $n$: \# Samples for computing $\underline{p_A}$.
    \item $\alpha$: Confidence parameters.
    \item $\mathcal{C}_f$: Cache storing information while certifying known prompts for input $x$. See text for details.
\end{itemize}

\textbf{Output}
\begin{itemize}
    \item Predicted class $c_A$ for input $x$, along with certified radius $R$ with $1 -\alpha$ confidence or ABSTAIN.
\end{itemize}
\hrule
\begin{algorithmic}[1]
\State $P$ $\leftarrow$ $f_p(prompt)$
\State $emb_{im}$ $\leftarrow$ $\mathcal{C}_f[emb]$.
\State $count0, count$ $\leftarrow$ \CountPrediction$(emb_{im}, P, n_0, n)$
\State $\hat{c}_A$ $\leftarrow$ top index in $counts0$.
\State $\underline{p_A}$  $\leftarrow$ \LowerConfBound$(counts[\hat{c}_A], n, 1 - \alpha)$.
\If{$\underline{p_A} > \frac{1}{2}$}
    \State \textbf{return} Class: $\hat{c}_A$, Radius: $\sigma \cdot \Phi^{-1}(p_A)$.
\Else
    \State \textbf{return} ABSTAIN.
\EndIf

\end{algorithmic}
\end{algorithm}

We use following information from the cache $\mathcal{C}_f$. Note that this cache has information specifically for the input $x$.
\begin{itemize}
    \item $\mathcal{C}_f[emb]$: Returns image embeddings for all the $n$ perturbations.
\end{itemize}

The algorithm also makes use of the following new functions:
\begin{itemize}
    \item \CountPrediction$(img\_emb\_arr, prompt\_emb, n_0, n)$: This is similar to the \SampleUnderNoise function, but receives precomputed image embeddings. It returns both the count using only $n_0$ samples adn complete $n$ samples.
\end{itemize}

This method gives us the exact same certificates as one would achieve using full forward passes through the classifier a la Cohen et al., \cite{cohen2019certified}. However, as the results indicate below, the caching trick enables \emph{two orders of magnitude faster} execution, since we no longer need to perform forward passes through the image encoder. The price we pay is the extra memory costs in storing all the embeddings, which we address next.

\subsection{A faster heuristic (MVN-OVC)}
In the OVC algorithm, we need to cache a large number (typically $100K$) of embeddings for each sample. This would consume hundreds of megabytes of memory for each sample for each noise setting, i.e., each value of $\sigma$. This is undesirable. 

To remedy this, we propose a heuristic approximation. Instead of saving the entire set of image embeddings, we fit a multivariate Gaussian (MVN) to the empirical distribution of the embeddings. While output of the image encoder need not be Gaussian, we are approximating it with mvn at a given point. This simple approximation saves a lot of storage space as now we only need to store only the mean ($\mu$) and covariance matrix ($\Sigma$) whose size is comparable to a single ImageNet image.

Once we have the MVN parameters we can easily sample from this distribution using standard Guassian samplers, and use Algo \ref{alg:ovc} for certification. Empirically, we observed that this heuristic gives a very good approximation of the certified radius. However, we noticed that for higher radius, sometimes the approximated radius exceeds the certified radius. We propose reducing the calculated $p_A$ by $1\%$ to get a lower estimate. 
As demonstrated in the scatter plots in our results, the certified radius is not exceeded for various settings, including different backbone models and datasets, suggesting that MVNs are effective for obtaining an approximate certification.

\textbf{MVN in logit space}: We notice that for open vocabulary  models like CLIP, there is a linear transformation from embedding space of images to logits. Further a Gaussian distribution remains Gaussian under linear transformation.
Thus, while certifying a novel prompt, we first transform the fitted Gaussian $\mathcal{N}(\mu, \Sigma)$ to the logit space. The transformed Gaussian is: $\mathcal{N}(P\mu, P\Sigma P^T).$. We present the MVN-OVC algorithm in Algo \ref{alg:approx_emb}.

\begin{algorithm}[htbp]
\caption{MVN-OVC$(f, \sigma, x, n_0, n, \alpha,\mathcal{C}_f)$}
\label{alg:approx_emb}
\textbf{Inputs:}
\begin{itemize}
    \item $f$: Given base vision-language model:
        \begin{itemize}
            \item $f_{im}$: Encodes image.
            \item $f_p$: Encodes prompt.
            \item $prompt$: Prompt for all classes.
        \end{itemize}
    \item $\sigma$: Std-dev of Gaussian noise used for certification.
    \item $x$: Input.
    \item $n_0$: \# Samples to predict the top class.
    \item $n$: \# Samples for computing $\underline{p_A}$.
    \item $\alpha$: Confidence parameters.
    \item $\mathcal{C}_f$: Cache storing information while certifying known prompts for input $x$. See text for details.
\end{itemize}

\textbf{Output}
\begin{itemize}
    \item Predicted class $c_A$ for input $x$, along with certified radius $R$ with $1 -\alpha$ confidence or ABSTAIN.
\end{itemize}
\hrule
\begin{algorithmic}[1]
\State $P$ $\leftarrow$ $f_p(prompt)$
\State $\mu, \Sigma$ $\leftarrow$ $\mathcal{C}_f[mvn]$.
\State $emb_{im}$ $\leftarrow$ \SampleFrom$(P\mu, P\Sigma P^T, n)$
\State $count0, count$ $\leftarrow$ \CountPrediction$(emb_{im}, P, n_0, n)$
\State $\hat{c}_A$ $\leftarrow$ top index in $counts0$.
\State $\underline{p_A}$  $\leftarrow$ \LowerConfBound$(counts[\hat{c}_A], n, 1 - \alpha)$.
\State $\underline{p_A}$  $\leftarrow$ $0.99 \times \underline{p_A}$
\If{$\underline{p_A} > \frac{1}{2}$}
    \State \textbf{return} Class: $\hat{c}_A$, Radius: $\sigma \cdot \Phi^{-1}(p_A)$.
\Else
    \State \textbf{return} ABSTAIN.
\EndIf

\end{algorithmic}
\end{algorithm}

We use following information from the cache $\mathcal{C}_f$. Note that this cache has information specifically for the input $x$.
\begin{itemize}
    \item $\mathcal{C}_f[mvn]$: Returns the mean ($\mu$) and Covariance ($\Sigma$) for the approximated multi-variate gaussian.
\end{itemize}

The algorithm also makes use of the following new functions:
\begin{itemize}
    \item SampleFrom$(\mu, \Sigma, n)$: Samples $n$ samples from the passed MVN parameters.
\end{itemize}

As the results indicate this is slightly faster than the theoretically sound OVC algorithm (Algo \ref{alg:ovc}). This is because, typically it is faster to sample from a mvn than to load the embeddings from the disk.


\section{Experiments}
We tested our method on CLIP and OpenCLIP using two datasets: ImageNet and CIFAR-10. The primary objective of this work is to expedite the standard RS certification process \cite{cohen2019certified} for zero-shot open vocabulary classifiers with novel prompts. It's important to note that the certificate's nature remains unchanged; that is, we neither enhance the certificates nor their accuracy. However, our method facilitates faster and more memory-efficient certification, especially for MVN-OVC. The key advantage of our approach lies in the accelerated certification speed, as detailed in Tables \ref{tab1} and \ref{tab:m_irs} in the main manuscript and Tables \ref{tab:speedups}, and \ref{tab:mem} in the appendix. For our experiments we used prompts from the official repository of CLIP at: \url{https://github.com/openai/CLIP}. It has 80 prompts for ImageNet and 18 for CIFAR-10. In line with the previous literature for ImageNet we calculated certificate for every $100^{th}$ sample and for CIFAR-10 we did it for every $20^{th}$ sample, unless otherwise stated. We also defer most of the results for OpenCLIP to the Appendix.

\begin{remark}
We wish to clarify that our method achieves performance gains in the certification of classifiers for novel prompts by utilizing pre-cached data from existing prompts. This approach is specifically applicable in zero-shot open-vocabulary classification scenarios. For a new data point with a single prompt, the computational cost aligns with that of Cohen et al. \cite{cohen2019certified}. The speedups are realized subsequently, i.e., when certifying novel prompts for the same data point.
\end{remark}

\begin{remark}
We observe that open-vocabulary models aren't restricted to particular datasets. These pretrained models include predefined transformations that should be applied to input data. In contrast to traditional certification methods that certify in the image space, our approach certifies in a normalized space, post-transformation. Our findings, detailed in Appendix \ref{clip_rob}, reveal that CLIP's robustness in native image space is quite limited.
\end{remark}

\subsection{CLIP certification}
\begin{figure}[!htbp]
    \centering
        \begin{subfigure}[b]{0.9\linewidth}
        \includegraphics[width=1.0\textwidth]{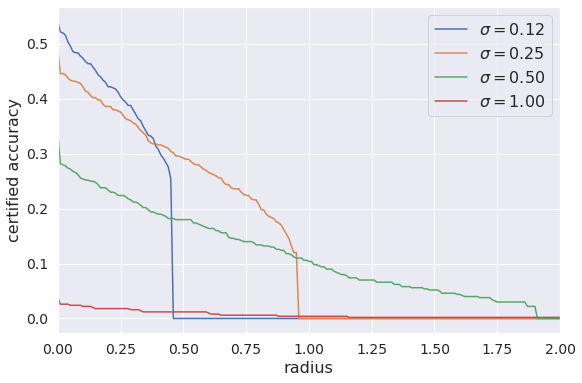}
        \caption{ImageNet}
        \end{subfigure}
        \vspace{1em} 
        \begin{subfigure}[b]{0.9\linewidth}
        \includegraphics[width=1.0\textwidth]{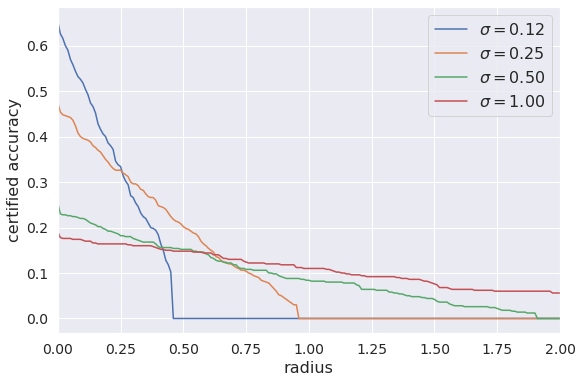}
        \caption{CIFAR-10}
        \end{subfigure}

    \caption{Certification of CLIP-RN50 model for various $\sigma$ for ImageNet and CIFAR-10 dataset.}
\label{fig:clip_cert}
\end{figure}

First we present the results of directly applying the standard RS certification method to CLIP. We present the result in Fig \ref{fig:clip_cert} for RN50 model.

We observer that for higher values of noise the certification accuracy drops considerably. This is expected as CLIP is trained on clean images, and we use the original pretrained model for all the settings.

\subsection{Modified-IRS}
\begin{figure}[!htbp]
    \centering
        \begin{subfigure}[b]{0.9\linewidth}
        \includegraphics[width=0.9\textwidth]{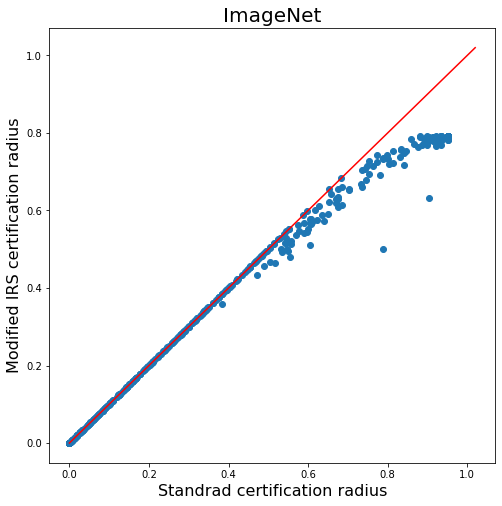}
        \caption{}
        \end{subfigure}
        \vspace{1em} 
        \begin{subfigure}[b]{0.9\linewidth}
        \includegraphics[width=0.9\textwidth]{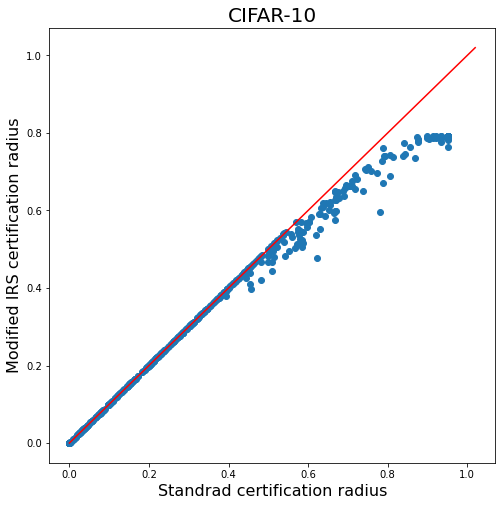}
        \caption{}
        \end{subfigure}

    \caption{Scatter plot comparing radius obtained using Modified-IRS and standard method (Cohen et al), for ImageNet \& CIFAR-10 for CLIP-RN50 with $\sigma=0.25$.}
\label{fig:m_irs}
\end{figure}

In our setting, for ImageNet, we randomly divided the $80$ prompts given in the CLIP offical repository, into $70$ known prompts and $10$ novel prompts. For CIFAR-10 out of $18$ prompts we used $15$ as known and $3$ as novel. We didn't used the average of all the prompt embeddings for prediction so that novel prompts could be kept novel. Using the Cohen et al., \cite{cohen2019certified} method, we computed the certification radius for all of them using $100K$ samples. We used the same seed for generating Gaussian noise while certifying all the prompts and saved the seeds. To test Modified-IRS we set $n_p = 10K$, i.e, for each input sample the algorithm uses $10K$ samples to find the prompt which is most similar in prediction to the novel prompt. If the difference in prediction is $< 1\%$, i.e., we set $\gamma = 0.01$, we use IRS method to compute radius, using the $p_A$ of the most similar known prompt, else we resort to Cohen et al. We show the result for a prompt (prompt id = 41) from the novel set in Fig \ref{fig:m_irs}. Note that, we are certifying the samples for the predicted top class, which need not be the correct class.

Modified-IRS method is able to considerably boost the speed up when compared to standard method, especially for lower values of $\sigma$. 
The result for various models and $\sigma$ has been presented in Table \ref{tab:m_irs}
\begin{table}
    \centering
    \begin{tabular}{cc | r r r r}
    \hline    
    Architecture & Dataset &  \multicolumn{4}{|c}{Speedup for $\sigma$} \\
    \hline
                 &       &  $0.12$ & $0.25$ & $0.50$ & $1.00$ \\
    \hline
         CLIP-RN50       & ImageNet & 1.68x   & 1.32x  & 1.02x  & 0.94x \\
                         & CIFAR-10  & 1.41x   & 1.20x  & 1.16x  & 1.02x \\
    \hline
      CLIP-ViT-B/32      & ImageNet & 2.18x   & 1.74x  & 1.32x  & 1.02x \\
                         & CIFAR-10  & 3.27x   & 2.38x  & 1.46x  & 1.08x \\
    \hline 
      Open-CLIP-ViT-B-32 & ImageNet & 2.49x   & 1.89x  & 1.33x  & 1.01x\\
                         & CIFAR-10  & 3.17x   & 2.04x  & 1.36x  & 1.07x \\
    
    \hline
    \end{tabular}
    \caption{Average speedup obtained for the test prompts using Modified-IRS for different architectures of CLIP for the two datasets}
    \label{tab:m_irs}
\end{table}
We note that, for higher level of noise (ie., high $\sigma$), the speedup is limited. In fact, when $\sigma = 1.0$, then for CLIP-RN50, for ImageNet, Modified-IRS takes slightly more time. This is because, as pointed in previous section, CLIP accuracy drops rapidly with noise. As a result, for a novel prompt, it becomes difficult to find an existing prompt which does not differ from it while making prediction. Thus for most of the input samples we need to resort to standard CERTIFY method (Step-5 of Modified-IRS Algo \ref{alg:m_IRS}). In the figure below, Fig \ref{fig:m_irs_agree}, we plot the fraction of input samples for which we were able to apply IRS and thus save compute time. We note that it monotonically decreases as $\sigma$ is increased.
\begin{figure}[!htbp]
    \centering
    \includegraphics[width=0.45\textwidth]{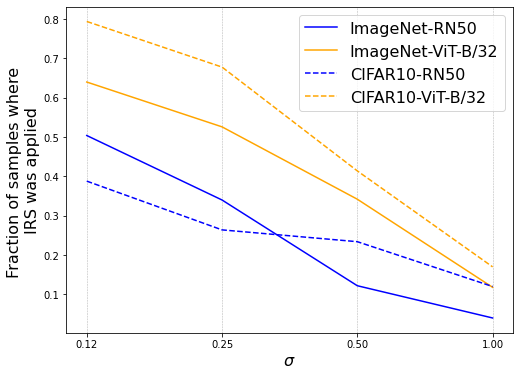}
    \caption{Plot showing fraction of input samples for which IRS was applied for CLIP.}
\label{fig:m_irs_agree}
\end{figure}

\begin{figure}[!htbp]
    \centering
    \includegraphics[width=0.49\textwidth]{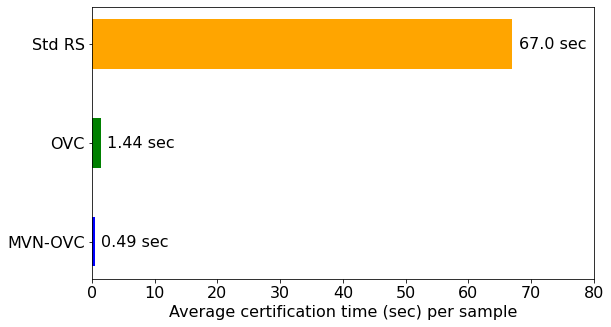}
    \caption{Comparison of Average Certification Time: OVC, MVN-OVC vs. Standard RS Method for CLIP-RN50.}
\label{fig:ovc_time}
\end{figure}
\subsection{OVC}
For this method we are pre-saving the image embeddings while certifying the known prompts. Since prompts are not utilized in calculating image embeddings we do not need to split the prompts in known and novel sets. 
For ImageNet, employing $100K$ perturbations and the standard RS certification method, each sample requires approximately 1 minute and 7 seconds for certification. However, when we save the image embeddings, the processing time is reduced significantly, by almost two orders of magnitude, with each sample now taking approximately 1.44 seconds. Please note that a substantial portion of the time is consumed during the loading of embeddings.

In Fig~\ref{fig:ovc_time}, we compare the time taken by OVC and MVN-OVC in comparison to the Standard RS method. Almost the entire duration for the Standard RS method is attributable to repeated passes of the input (with added Gaussian noise) through the model. In contrast, for OVC, the primary time expenditure is associated with reading the embeddings from disk. Therefore, these times can be independently adjusted depending on system configurations.

\begin{figure}[!htbp]
    \centering
    \includegraphics[width=0.45\textwidth]{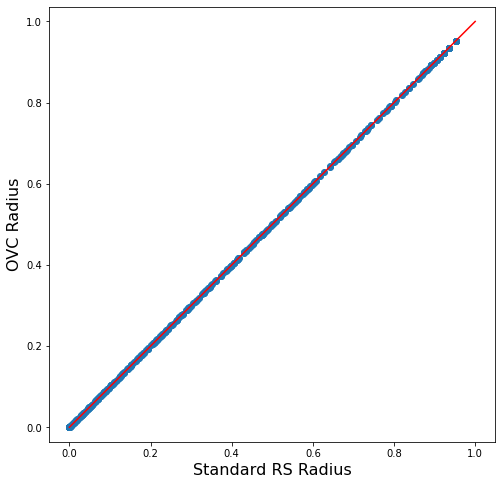}
    \caption{Scatter plot comparing radius obtained using OVC and standard RS method, for CLIP-RN50 with $\sigma=0.25$ for ImageNet for a random prompt.}
\label{fig:ovc}
\end{figure}

This method returns the same radius as obtained by the standard method, as shown in Fig \ref{fig:ovc}.

\subsection{MVN-OVC}
\begin{figure}[!htbp]
\centering
\begin{subfigure}{0.45\textwidth}  
        \centering
        \includegraphics[width=\textwidth]{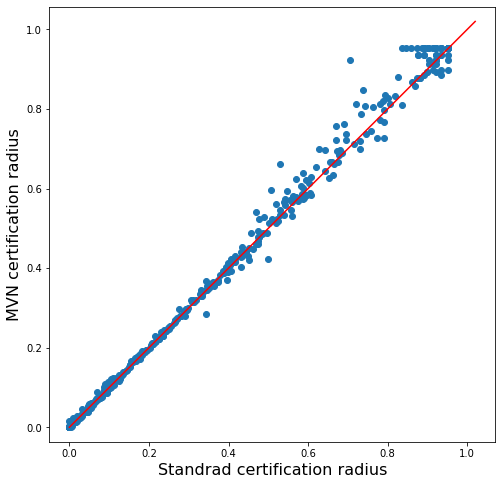}
        \caption{ImageNet}
    \end{subfigure}%
    
    \begin{subfigure}{0.45\textwidth}  
        \centering
        \includegraphics[width=\textwidth]{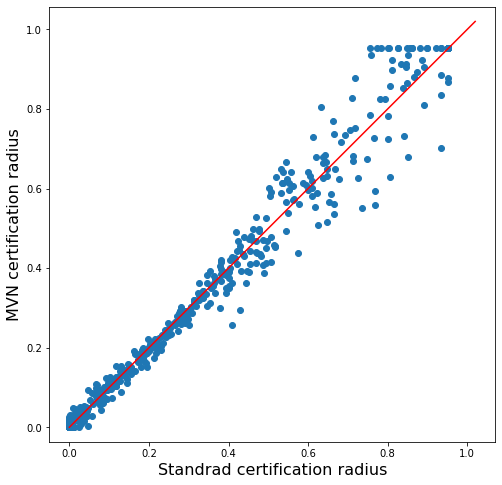}
        \caption{CIFAR-10}
    \end{subfigure}%
    \caption{Scatter plot comparing the certified radius as obtained by MVN and standard RS method for CLIP-RN50 with $\sigma = 0.25$ for ImageNet and CIFAR-10 datasets.}
\label{fig:app_ovc}
\end{figure}

Here instead of storing the entire $100K$ embeddings, we approximate it via a mvn and store the parameters $\mu$ and $\Sigma$. For a novel prompt, $P$ we transform the mvn to logit space and directly sample the logits from $\mathcal{N}(P\mu, P\Sigma P^T, n)$. While this gives approximately correct radius, for larger radius it sometimes overshoots the actual certification radius as calculated using the standard way. This has been illustrated in Fig \ref{fig:app_ovc} for both CIFAR-10 and ImageNet.

\begin{figure}[!htbp]
\centering
\begin{subfigure}{0.43\textwidth}  
        \centering
        \includegraphics[width=\textwidth]{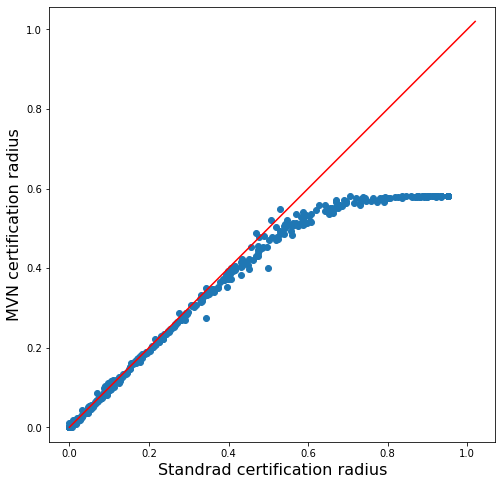}
        \caption{ImageNet}
    \end{subfigure}%
    
    \begin{subfigure}{0.43\textwidth}  
        \centering
        \includegraphics[width=\textwidth]{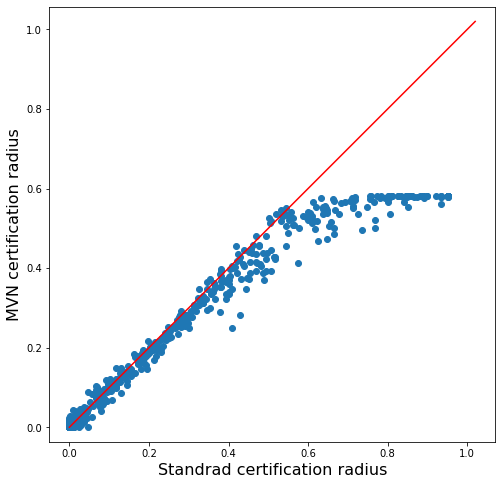}
        \caption{CIFAR-10}
    \end{subfigure}%
    \caption{Scatter plot comparing the certified radius as obtained by MVN-OVC after reducing the estimated $p_A$ by $1\%$ and standard RS method for CLIP-RN50 with $\sigma = 0.25$ for ImageNet and CIFAR-10 datasets.}
\label{fig:app_ovc_red}
\end{figure}

For a certified radius, the certification process shall never overestimate the radius. A simple way to fix this is to find a bound on the error and reduce the estimated $p_A$ by that bound. Finding a bound analytically does not seem tractable for such a high dimensional data. Thus instead, we tested our method empirically by reducing the estimated $p_A$ by small fraction. We empirically found that by reducing the probability by as little as $1\%$, the estimated certification radius as calculated by MVN-OVC does not exceed the radius obtained using standard method. This however, as expected, caps the radius at higher values as $p_A$ will never exceed $0.99$. We present the scatter plots in Fig~\ref{fig:app_ovc_red}. Results for OpenCLIP and more backbone architecture has been deferred to appendix.

\subsection{Speedup Breakdown}
Different approaches yield varying degrees of speedup. For Modified-IRS, speed is gained by reducing the number of samples needed for certification, dependent on factors like the noise level (i.e., the value of $\sigma$), the presence of a closest prompt, and the dataset. However, as shown in Table \ref{tab:m_irs}, the gain is relatively modest.

The primary time consumption in the standard RS algorithm \cite{cohen2019certified} is due to multiple forward passes. We found that this can be mitigated by the implementation of Cached-OVC and subsequently MVN-OVC, achieving significant speed increases for novel prompts compared to standard RS.

For Cached-OVC, the majority of time is spent loading the cached embeddings, as a substantial amount of data must be transferred from disk to GPU. For instance, loading 100K CLIP-RN50 embeddings in Cached-OVC takes about 1.4 seconds, while loading the MVN parameters in MVN-OVC takes less than 0.2 seconds, making MVN-OVC roughly three times faster than Cached-OVC.

\textbf{Memory Performance for OVC Methods}
Modified-IRS significantly reduces memory usage, as it eliminates the need to store embeddings, resulting in a considerably smaller memory footprint. Specifically, it requires approximately 40.1MB of storage per prompt.

We present a comparison of both speed and memory utilization for Cached-OVC and MVN-OVC in Table \ref{tab:mem}.

\begin{table}[!htbp]
    \centering
    \begin{tabular}{l | r r}
    \hline    
                  & Cached-OVC       & OVC-MVN\\
    \hline
    Memory for 1 image  & 204.8 MB    & 8.4 MB \\
    For 500 images & 100 GB & 4.1 GB \\
    Speed-up   & $\sim$ 46x    & $\sim$ 137x \\
    \hline
    \end{tabular}
    \caption{Comparing speed and memory usage for Cached-OVC and OVC-MVN for CLIP-RN50 when certified using 100K samples per input.}
    \label{tab:mem}
\end{table}

\section{Discussion and Conclusion}
In this paper, we present and empirically validate a framework designed for certifying zero-shot vision-language classifiers through randomized smoothing. A compelling attribute of these classifiers lies in their flexibility: users have the freedom to create and employ novel prompts for classification at the inference stage.
To address this scenario, we have developed specialized certification techniques. Specifically, our methods expedite the certification process for novel prompts, drawing upon pre-existing certifications and related metadata for known prompts.

Our first proposed method, Modified-IRS, searches for a prompt whose prediction is most similar to that of the novel prompt for a given input. Upon finding such a prompt, it quickly generates a certificate for the novel prompt. We observed that this method significantly speeds up the certification process for various values of $\sigma$.


We then introduced the Open Vocabulary Certification (OVC) algorithm, which leverages the fact that for models like CLIP, the image embeddings remain constant for novel prompts. Consequently, we cache these embeddings to expedite the certification process. However, this approach increases storage demands, as RS requires thousands of input perturbations. To address this, we employ a heuristic multivariate normal (MVN) approximation of the embedding space for each input. Given the linear relationship between the logit and embedding spaces via prompt embeddings, we can quickly derive the approximated distribution of the logit space. This results in further speedup, as we can sample from the MVN much faster than loading embeddings from disk. While the MVN provides certification radii remarkably close to standard RS methods, it sometimes slightly overshoots the prediction for larger radii. We successfully mitigated this by reducing the underlying probability of the top class by a small amount, i.e., $1\%$.

There are several avenues we would like to explore further in the future. Quantifying the error in the MVN approximation could be invaluable, as it would allow us to achieve fast probabilistic certification. Additionally, we currently have to approximate the MVN separately for each value of $\sigma$. We plan to investigate whether we can obtain embeddings for different $\sigma$ values using a single distribution.

We also observed that while these zero-shot vision-language models offer natural accuracy comparable to traditional models, they lack robustness, particularly in the original image space. This is expected since these models are not exposed to adversarial or noisy images during training. Although retraining these models would be costly, exploring alternative solutions, such as image pre-processing, could be a valuable avenue for improving their robustness.

{\small
\bibliographystyle{plain}
\bibliography{main}
}

\clearpage
\appendix

\subsection{Modified IRS agreement}
\begin{figure}[!htbp]
    \centering
    \includegraphics[width=0.45\textwidth]{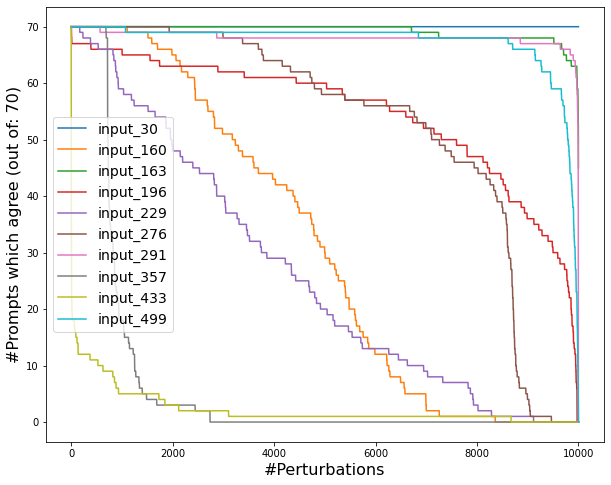}
    \caption{Agreement of individual samples for a novel prompt wrt known prompts. Please see the text for details.}
\label{fig:inp_agree}
\end{figure}

In Fig~\ref{fig:simPrompt} we showed that, for CLIP with RN50 backbone, when $\sigma=0.25$, for a novel prompt, for about $30\%$ ImageNet samples, difference in prediction with at least one existing prompt is small enough to apply IRS. In Fig~\ref{fig:inp_agree} we show the same agreement for individual samples. Specifically for the novel prompt, we randomly picked 10 inputs and plotted the agreement in their prediction with existing 70 prompts. We notice that the agreement varies wildly for different inputs. While, for some inputs (like input 499) the predictions for all the prompts matches perfectly for all the 10K perturbations, for many other inputs it drops fairly quickly.

\subsection{More results on Certification}
\begin{figure}
    \centering
    \begin{subfigure}{0.40\textwidth}
        \includegraphics[width=\linewidth]{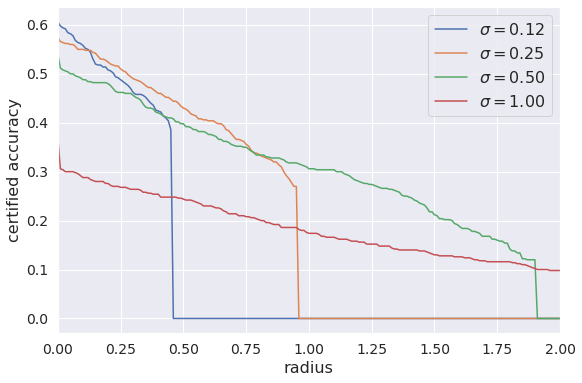}
        \caption{CLIP-ViT-B/32 for ImageNet}
        \label{fig:c_cert_sub1}
    \end{subfigure}
    \begin{subfigure}{0.40\textwidth}
        \includegraphics[width=\linewidth]{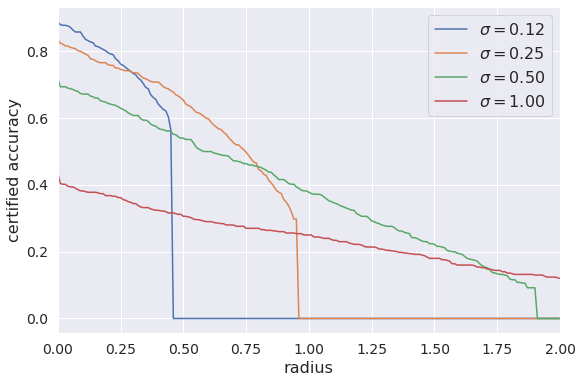}
        \caption{CLIP-ViT-B/32 for CIFAR-10}
        \label{fig:c_cert_sub2}
    \end{subfigure}

    \begin{subfigure}{0.40\textwidth}
        \includegraphics[width=\linewidth]{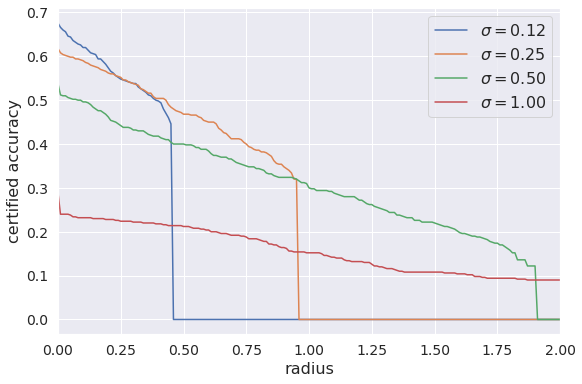}
        \caption{OpenCLIP-ViT-B-32 for ImageNet}
        \label{fig:c_cert_sub3}
    \end{subfigure}
    \begin{subfigure}{0.40\textwidth}
        \includegraphics[width=\linewidth]{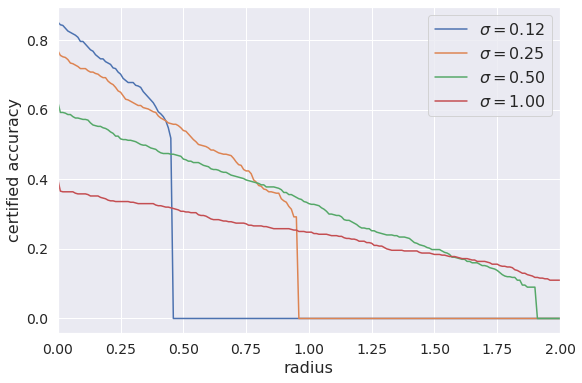}
        \caption{OpenCLIP-ViT-B-32 for CIFAR-10}
        \label{fig:c_cert_sub4}
    \end{subfigure}
    \caption{Certification for CLIP and OpenCLIP with respective ViT backbone for ImageNet and CIFAR-10 datasets.}
    \label{fig:c_cert}
\end{figure}

We present the certification results, obtained using RS, for CLIP with a ViT-B/32 backbone and OpenCLIP with ViT-B-32, as shown in Fig~\ref{fig:c_cert}. Our observations indicate that the ViT backbone delivers superior accuracy and robustness when compared to RN50. Notably, the results for CLIP and OpenCLIP are closely aligned, with CLIP demonstrating a slight performance edge over OpenCLIP.

\subsection{More results for MVN-OVC}

\begin{figure*}
    \centering
    \begin{subfigure}{0.45\textwidth}
        \includegraphics[width=\linewidth]{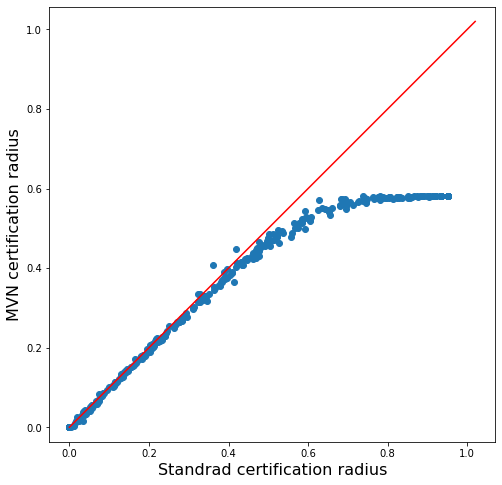}
        \caption{CLIP-ViT-B/32 for ImageNet}
        \label{fig:mvn_ovc_cert_sub1}
    \end{subfigure}
    \begin{subfigure}{0.45\textwidth}
        \includegraphics[width=\linewidth]{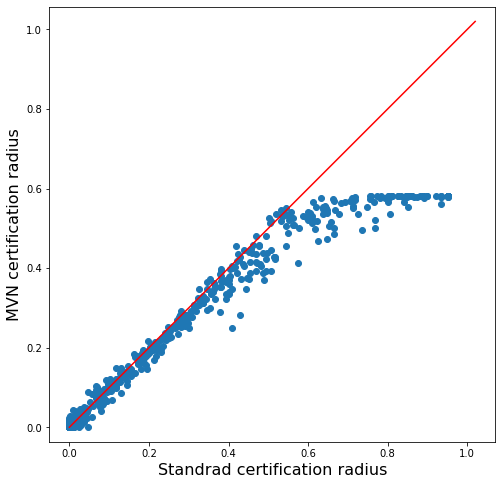}
        \caption{CLIP-ViT-B/32 for CIFAR-10}
        \label{fig:mvn_ovc_cert_sub2}
    \end{subfigure}

    \begin{subfigure}{0.45\textwidth}
        \includegraphics[width=\linewidth]{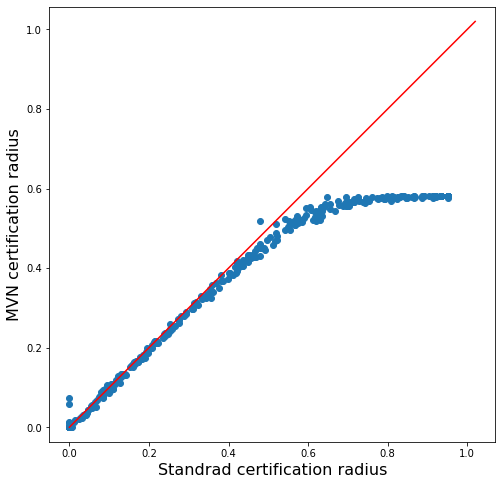}
        \caption{OpenCLIP-ViT-B-32 for ImageNet}
        \label{fig:mvn_ovc_cert_sub3}
    \end{subfigure}
    \begin{subfigure}{0.45\textwidth}
        \includegraphics[width=\linewidth]{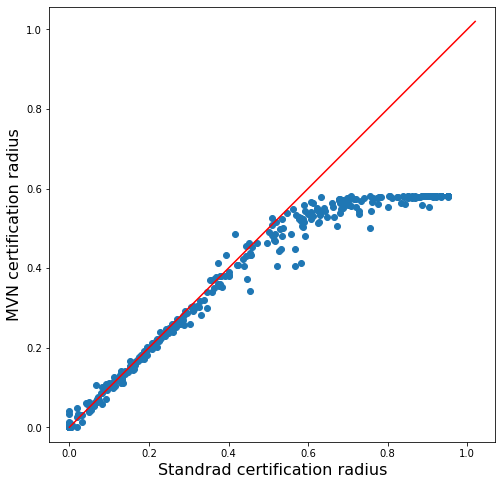}
        \caption{OpenCLIP-ViT-B-32 for CIFAR-10}
        \label{fig:mvn_ovc_cert_sub4}
    \end{subfigure}
    \caption{Scatter plots comparing certification results for MVN-OVC method vs RS method for CLIP and OpenCLIP with specified ViT backbone for ImageNet and CIFAR-10 datasets for $\sigma=0.25$.}
    \label{fig:mvn_ovc_cert}
\end{figure*}

\begin{figure*}
    \centering 

    \begin{subfigure}{0.40\textwidth}
        \includegraphics[width=\linewidth]{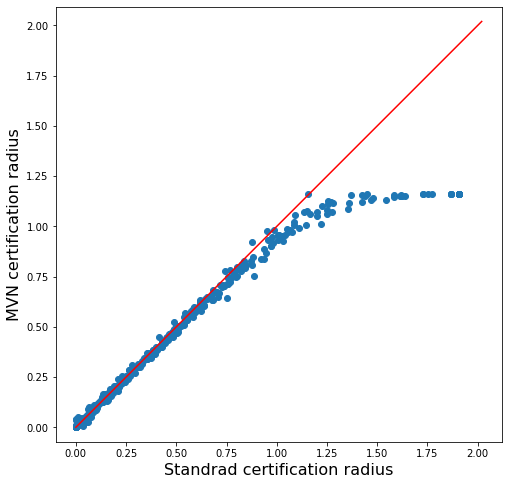}
        \caption{CLIP-RS50 for ImageNet}
        \label{fig:mvn_ovc_cert_050_sub1}
    \end{subfigure}
    \begin{subfigure}{0.40\textwidth}
        \includegraphics[width=\linewidth]{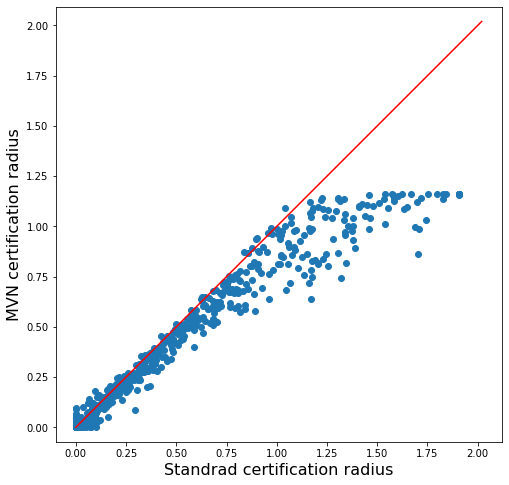}
        \caption{CLIP-RS50 for CIFAR-10}
        \label{fig:mvn_ovc_cert_050_sub2}
    \end{subfigure}
    
    \begin{subfigure}{0.40\textwidth}
        \includegraphics[width=\linewidth]{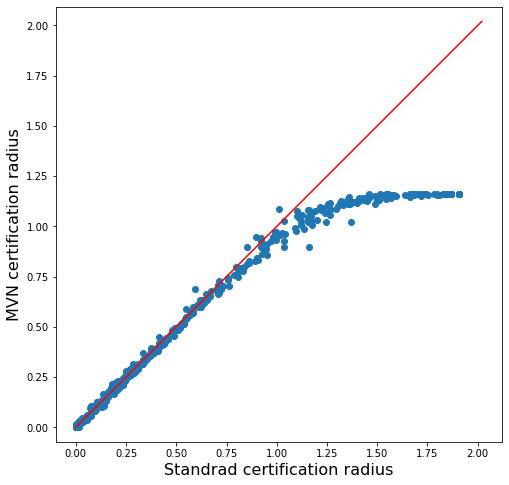}
        \caption{CLIP-ViT-B/32 for ImageNet}
        \label{fig:mvn_ovc_cert_050_sub3}
    \end{subfigure}
    \begin{subfigure}{0.40\textwidth}
        \includegraphics[width=\linewidth]{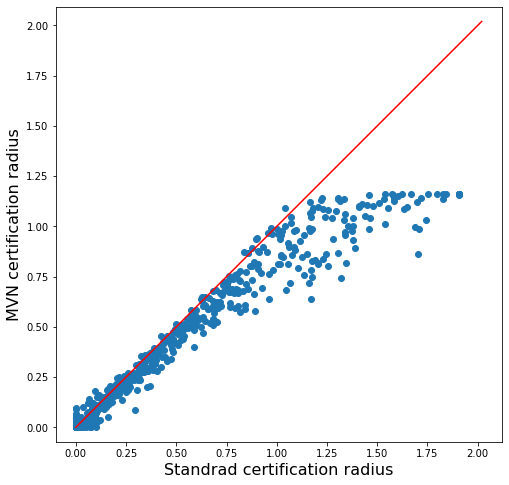}
        \caption{CLIP-ViT-B/32 for CIFAR-10}
        \label{fig:mvn_ovc_cert_050_sub4}
    \end{subfigure}

    \begin{subfigure}{0.40\textwidth}
        \includegraphics[width=\linewidth]{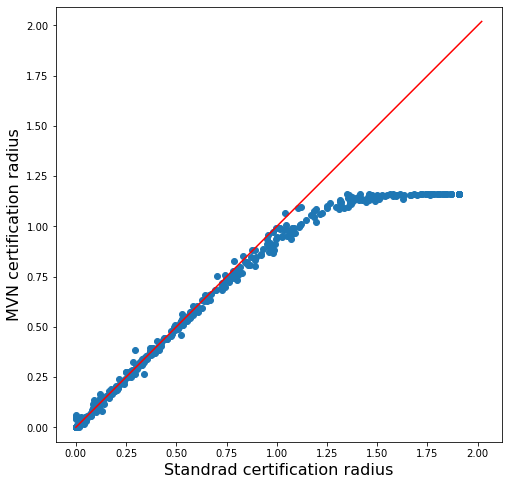}
        \caption{OpenCLIP-ViT-B-32 for ImageNet}
        \label{fig:mvn_ovc_cert_050_sub5}
    \end{subfigure}
    \begin{subfigure}{0.40\textwidth}
        \includegraphics[width=\linewidth]{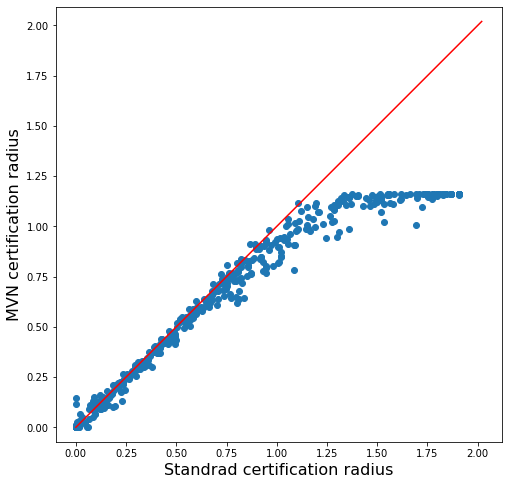}
        \caption{OpenCLIP-ViT-B-32 for CIFAR-10}
        \label{fig:mvn_ovc_cert_050_sub6}
    \end{subfigure}
    \caption{Scatter plots comparing certification results for MVN-OVC method vs RS method for CLIP and OpenCLIP with specified backbone for ImageNet and CIFAR-10 datasets for $\sigma=0.50$.}
    \label{fig:mvn_ovc_cert_050}
\end{figure*}

In Fig~\ref{fig:mvn_ovc_cert}, we present the certification results, comparing certification obtained using MVN-OVC and the standard RS method for CLIP with a VIT-B/32 backbone and OpenCLIP with a VIT-B-32 backbone, using $\sigma=0.25$. Fig~\ref{fig:mvn_ovc_cert_050} showcases similar scatter plots for $\sigma=0.50$, focusing on CLIP and OpenCLIP with the  backbone specified in the caption. Throughout all settings, we utilized the first prompt. Our observation reveals a consistent trend: the MVN-OVC method, with $1\%$ reduction in $p_A$, consistently underestimates the radius in comparison to the standard RS method.

\clearpage
\clearpage  

\subsection{Speedup for Different CLIP-Backbone Architectures}
The time required for a forward pass varies with model size. As models become larger, forward passes tend to take longer, thus increasing the relative speedup. We detail the time taken to obtain 100K predictions and the speedup for different CLIP-backbone architectures in Cached-OVC and MVN-OVC in Table \ref{tab:speedups} below.

\begin{table}[!h]
    \centering
    \begin{tabular}{c | r r r r}
    \hline    
    CLIP         & Emb' & Time to get & \multicolumn{1}{c}{Cached-OVC} & \multicolumn{1}{c}{MVN-OVC} \\
    Architecture & dim $\#$      & 100K emb'   & \multicolumn{1}{c}{Speedup}     & \multicolumn{1}{c}{Speedup} \\
    
    \hline
    RN50     & 1024 & 67.0  & 46x   & 136x\\     
    RN101    & 512  & 98.5  & 187x  & 581x\\
    RN50x4   & 640  & 211.5 & 341x  & 884x\\
    RN50x16*  & 768  & 578.5 & 1012x & 2494x\\
    ViT-B/32 & 512  & 35.0  & 70x & 217x\\
    ViT-B/16 & 512  & 146.5 & 286x & 975x\\
    
    \hline
    \end{tabular}
    \caption{Average (approximate) speedup obtained for various back bone architecture for CLIP. The speedup is almost identical for both ImageNet and CIFAR-10 datasets.}
    * For RN50x16, we needed to reduce the batch size from 400 to 200 for certifying using standard RS method.
    \label{tab:speedups}
\end{table}

The speedups are approximations (and are conservative), influenced by various disk-reading factors.The data load correlates with the embedding size. For three backbone architectures with identical embedding dimensions (512), namely RN101, ViT-B/32, and ViT-B/16, the speedups correspond to the duration needed to acquire 100K embeddings. All speedup measurements are conducted using an Nvidia GeForce RTX 2080 Ti graphics card and a Seagate Expansion Desktop 10TB External Hard Drive HDD.

\subsection{CLIP robustness}
\label{clip_rob}
\begin{figure}[!htbp]
    \centering
    \includegraphics[width=0.40\textwidth]{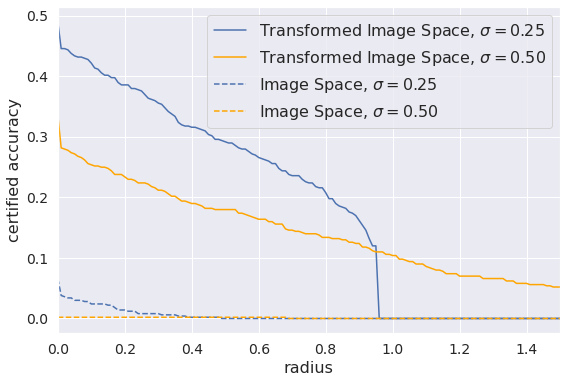}
    \caption{Comparing certification for CLIP for RN50 backbone in Image Space and transformed Image Space on ImageNet dataset.}
\label{fig:i_v_t_space}
\end{figure}

Throughout the certification process, we first transform the image using the transformation accompanied by these vision-language models. This is slightly different than the standard RS certification process, which certifies the model in the native image space. We observed that certification of CLIP in native image space is very limited as shown in Fig~\ref{fig:i_v_t_space}. 

\begin{figure}[!htbp]
    \centering
    \includegraphics[width=0.40\textwidth]{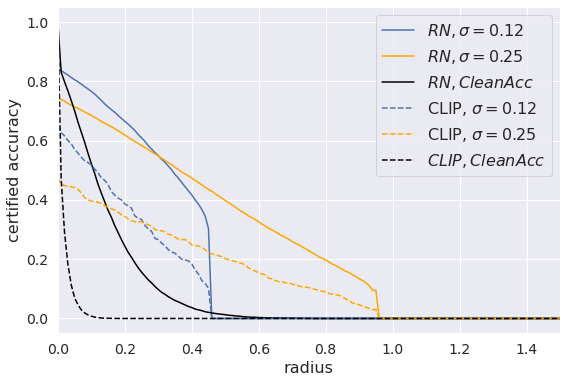}
    \caption{Comparing certification for CLIP, with ResNet-101 for various $\sigma$ on CIFAR-10 dataset}
\label{fig:CLIP_vs_RN}
\end{figure}

We also compare the certification of CLIP with ResNet models on CIFAR-10 testset as shown in Fig~\ref{fig:CLIP_vs_RN}. Here ResNet models are certified in the image space while CLIP is certified in the transformed image space. We observe that while natural accuracy for CLIP is on par with the ResNet-101 models, it has a lower robustness. Clean accuracy radius for both the models have been achieved using Deepfool \cite{moosavi2016deepfool} method as implemented by Foolbox\cite{rauber2017foolboxnative}. The $\ell_2$ radius calculated for CLIP has also been calculated in the transformed space where the image is scaled to $224 \times 224$ pixels. The corresponding radius in the original $32 \times 32$ pixels is considerably smaller.

\subsection{IRS Algorithm Overview} \label{app:IRS}

\begin{algorithm}[htbp]
\caption{IRS algorithm: Certification with cache}
\label{alg:IRS}
\textbf{Inputs:} $f^p$: DNN obtained from approximating $f$ , $\sigma$: standard deviation, $x$: input to the DNN, $n_p$ : number of Gaussian samples used for certification, $C_f$ : stores the information to be reused from certification of $f$ , $\alpha$ and $\alpha_\zeta$ : confidence parameters, $\gamma$: threshold hyperparameter to switch between estimation methods
\begin{algorithmic}[1]
\Function{CertifyIRS}{$f^p, \sigma, x, n_p, C_f, \alpha, \alpha_\zeta, \gamma$}
\State $\hat{c}_A$ $\leftarrow$ top index in $C_f[x]$
\State $\underline{p_A}$ $\leftarrow$ lower confidence $f$ from  $C_f[x]$
\If{$\underline{p_A} < \gamma$}
    \State $\zeta_x$ $\leftarrow$ EstimateZeta($f^p, \sigma, x, n_p, C_f, \alpha_\zeta$)
    \If{$\underline{p_A} - \zeta_x > \frac{1}{2}$}
        \State \textbf{return} prediction $\hat{c}_A$ and radius $\sigma\Phi^{-1}(\underline{p_A} - \zeta_x )$
    \EndIf
\Else
    \State $counts \gets$ SampleUnderNoise($f^p, x, n_p, \sigma$)
        \State $p'_A \gets$ LowerConfidenceBound($counts[\hat{c}_A], n_p, 1 - (\alpha + \alpha_\zeta)$)
        \If{$p'_A > \frac{1}{2}$}
            \State \textbf{return} prediction $\hat{c}_A$ and radius $\sigma \Phi^{-1}(p'_A)$
        \EndIf
\EndIf
\State \textbf{return} ABSTAIN

\EndFunction
\end{algorithmic}
\end{algorithm}

\begin{algorithm}
\caption{Estimate $\zeta_x$}
\label{alg:ZetaEst}
\textbf{Inputs:} $f^p$: DNN obtained from approximating $f$, $\sigma$: standard deviation, $x$: input to the DNN, $n_p$: number of Gaussian samples used for estimating $\zeta_x$, $C_f$: stores the information to be reused from certification of $f$, $\alpha_c$: confidence parameter
\textbf{Output:} Estimated value of $\zeta_x$
\begin{algorithmic}[1]
\Function{EstimateZeta}{$f^p, \sigma, x, n_p, C_f, \alpha_c$}
    \State $n_\Delta \gets 0$
    \State $seeds \gets$ seeds for original samples from $C_f[x]$
    \State $predictions \gets f$’s predictions on samples from $C_f[x]$
    \For{$i \gets 1, n_p$}
        \State $\epsilon \sim \mathcal{N}(0, \sigma^2)$ using $seeds[i]$
        \State $c_f \gets predictions[i]$
        \State $c_{fp} \gets f^p(x + \epsilon)$
        \State $n_\Delta \gets n_\Delta + \mathbb{I}(c_f \neq c_{fp})$
    \EndFor
    \State \textbf{return} UpperConfidenceBound($n_\Delta, n_p, 1 - \alpha_\zeta$)
\EndFunction
\end{algorithmic}
\end{algorithm}

\begin{figure*}[!htbp]
    \centering
    \includegraphics[width=0.9\textwidth]{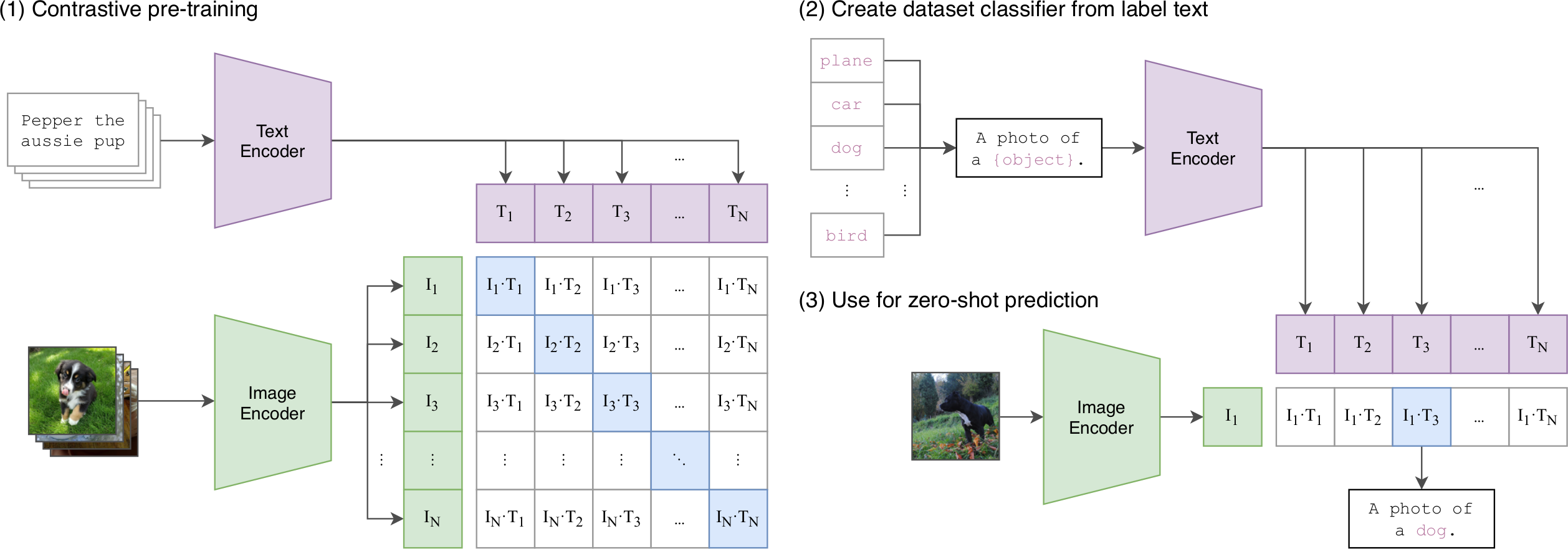}
    \caption{Training and Prediction Process of CLIP Visualized. Image sourced from the CLIP paper \cite{radford2021learning} to provide a comprehensive overview.}
\label{fig:clip}
\end{figure*}

Here, we detail the Incremental Randomized Smoothing (IRS) algorithm, as originally outlined in \cite{ugare2023incremental}. For ease of reference, Algorithm \ref{alg:IRS} delineates the core IRS procedure, while Algorithm \ref{alg:ZetaEst} describes the associated subroutine responsible for calculating the error difference.

\subsection{Overview of CLIP} \label{sec:clipOverview}
This section offers an overview of CLIP \cite{radford2021learning}, a zero-shot, open vocabulary classifier introduced by OpenAI in 2021. CLIP revolutionized image classification by training on a broad array of internet-sourced image-caption pairs, unlike traditional classifiers limited to specific datasets.

To assemble the training dataset, the authors utilized 500,000 queries, including high-frequency Wikipedia words and bi-grams, each capped at 20,000 (image, text) pairs, resulting in 400 million pairs overall. The images were assembled from various open sources. The text vocabulary comprised 49,152 words, and text length per image was limited to 76.

\begin{figure}[!htbp]
    \centering
    \includegraphics[width=0.45\textwidth]{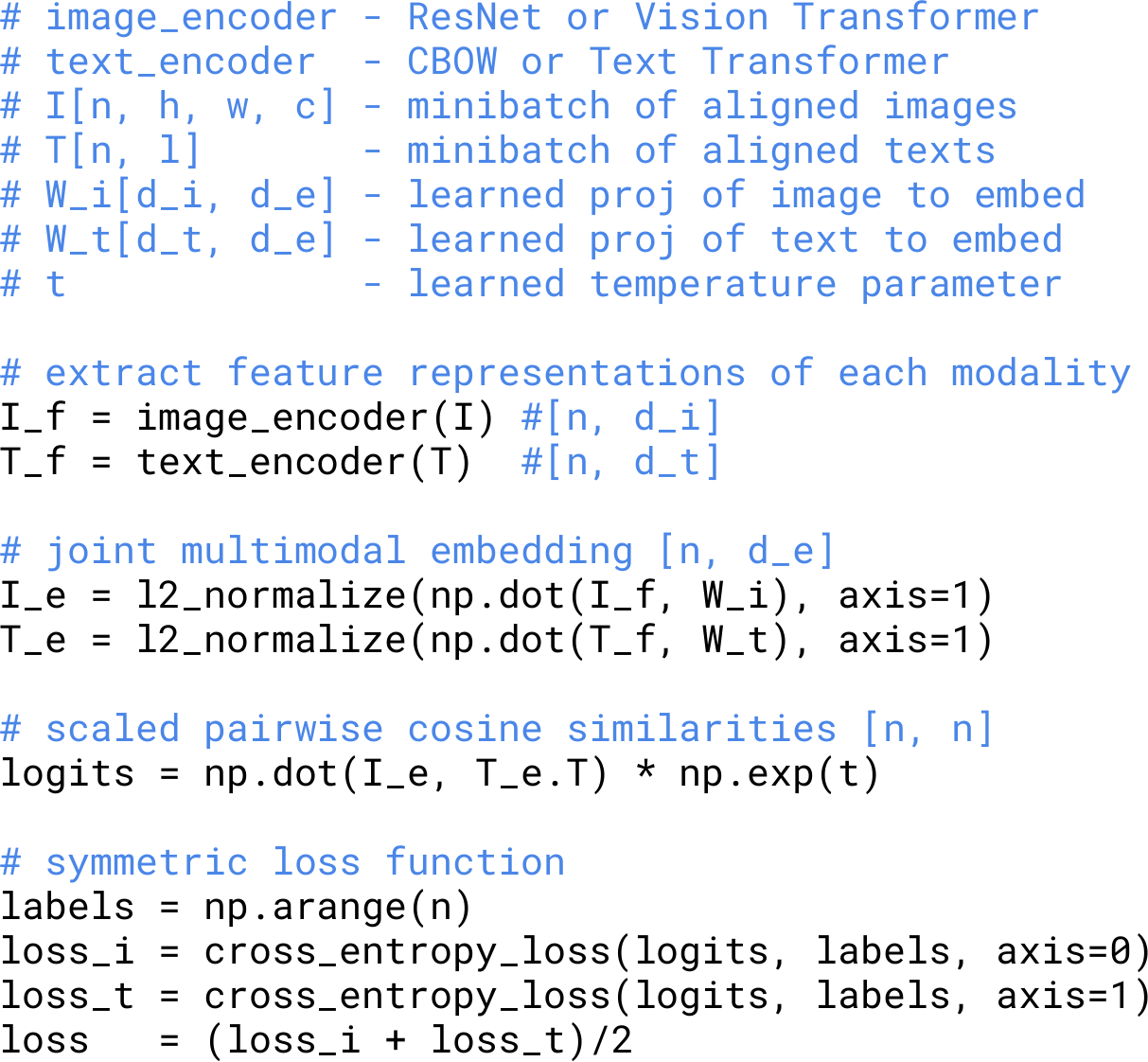}
    \caption{Pseudocode Depicting CLIP's Loss Function Implementation. Sourced from the CLIP paper \cite{radford2021learning} for illustrative purposes.}
\label{fig:CLIP_pseudo}
\end{figure}

Figure \ref{fig:clip} illustrates CLIP's training and prediction process, sourced from cite{radford2021learning}. The training involves encoding images and texts separately, aiming to align the encodings (embeddings) for each pair. With a batch of $N$ pairs, the goal is to distinguish the $N$ correct from $N^2 - N$ incorrect pairings, using cosine similarity in the loss function. The pseudo code, borrowed form the original paper is presented in Figure \ref{fig:CLIP_pseudo}. It creates logit values, by calculating cosine similarity, for each possible $N^2$ pairs for a batch. The loss is calculated using cross entropy, both for image-to-text and text-to-image directions. The final loss is the average of the two, providing a symmetric loss that ensures the model learns to align both image and text embeddings effectively. A large batch size of 32,768 was used. For comprehensive details on the encoders and training, refer to the original CLIP paper.

In application, CLIP serves as a zero-shot classifier. For classification, text prompts representing potential classes are created. For a dataset it could be text containing names of each class. An example of this is given in Figure \ref{fig:clip}. An image is classified based on the highest cosine similarity between its embedding and the class prompt's embedding. Multiple prompts per class are often averaged for classification. Sample prompts for various datasets are available in CLIP's official repository \url{https://github.com/openai/CLIP/tree/main}.

\end{document}